\documentclass{article}
\usepackage[english]{babel}

\usepackage[preprint]{neurips_2024}

\usepackage[utf8]{inputenc} 
\usepackage[T1]{fontenc}    
\usepackage{hyperref}       
\usepackage{url}            
\usepackage{booktabs}       
\usepackage{amsfonts}       
\usepackage{amsmath}
\usepackage{nicefrac}       
\usepackage{microtype}      
\usepackage{xcolor}         
\usepackage{subfiles}
\usepackage{comment}
\usepackage{enumitem}
\usepackage[pdftex]{graphicx} 
\usepackage[export]{adjustbox}
\usepackage[rightcaption]{sidecap}
\usepackage{wrapfig}
\usepackage{tcolorbox}
\usepackage{tablefootnote}
\usepackage[autostyle=true]{csquotes} 
\usepackage{float} 
\usepackage{colortbl}
\definecolor{Gray}{gray}{0.95}
\usepackage[font=small]{caption}
\usepackage[labelformat=empty, position=top]{subcaption}
\usepackage[export]{adjustbox}
\usepackage[ruled,vlined]{algorithm2e}
\usepackage{placeins}
\usepackage{multirow}
\usepackage{cleveref}

\definecolor{bluecite}{HTML}{0875b7}

\newtcolorbox{mybox}[1]{colback=bluecite!5!white,colframe=bluecite!75!black,fonttitle=\bfseries,title=#1}

\hypersetup{
  colorlinks=true,
  citecolor=bluecite,
  linkcolor=magenta
}

\title{Characterizing MARL for Energy Control: A Multi-KPI Benchmark on the CityLearn Environment

}

\author{%
  Aymen Khouja \\
  InstaDeep\\
  \texttt{a.khouja@instadeep.com} \\
  \And 
  Imen Jendoubi\\
  InstaDeep \\
  \And
  Oumayma Mahjoub \\
  InstaDeep \\
  \AND
   Oussama Mahfoudhi \\
  InstaDeep \\
  \And
    Ruan De Kock\\
  InstaDeep \\
  \And
  Siddarth Singh \\
  InstaDeep \\
  \And
  Claude Formanek \\
  InstaDeep \\
}

\begin{document}

\maketitle

\begin{abstract}
The optimization of urban energy systems is crucial for the advancement of sustainable and resilient smart cities, which are becoming increasingly complex with multiple decision-making units. To address scalability and coordination concerns, Multi-Agent Reinforcement Learning (MARL) is a promising solution. This paper addresses the imperative need for comprehensive and reliable benchmarking of MARL algorithms on energy management tasks. CityLearn is used as a case study environment because it realistically simulates urban energy systems, incorporates multiple storage systems, and utilizes renewable energy sources. By doing so, our work sets a new standard for evaluation, conducting a comparative study across multiple key performance indicators (KPIs). This approach illuminates the key strengths and weaknesses of various algorithms, moving beyond traditional KPI averaging which often masks critical insights. Our experiments utilize widely accepted baselines such as Proximal Policy Optimization (PPO) and Soft Actor Critic (SAC), and encompass diverse training schemes including Decentralized Training with Decentralized Execution (DTDE) and Centralized Training with Decentralized Execution (CTDE) approaches and different neural network architectures. Our work also proposes novel KPIs that tackle real world implementation challenges such as individual building contribution and battery storage lifetime. Our findings show that DTDE consistently outperforms CTDE in both average and worst-case performance. Additionally, temporal dependency learning improved control on memory dependent KPIs such as ramping and battery usage, contributing to more sustainable battery operation. Results also reveal robustness to agent or resource removal, highlighting both the resilience and decentralizability of the learned policies.
\end{abstract}
\section{Introduction}
\vspace{-0.25cm}
Electricity though indispensable, remains a significant contributor to carbon emissions. The integration of distributed energy resources (DERs), such as photovoltaic systems, wind turbines, and energy storage systems, is essential for reducing reliance on fossil fuels and mitigating environmental impacts \citep{9376271}. However, managing Distributed Energy Resources (DERs) at scale presents complex challenges, including the need for highly flexible and responsive energy management systems capable of fast real time operation. To address this, Demand Response (DR) mechanisms, which adjust electricity use based on supply conditions or price signals, are key to improving system flexibility and ensuring real-time balance between demand and supply. However, traditional approaches, often reliant on manual intervention, struggle to address the dynamic and time sensitive demands of modern energy systems \citep{dr_events_in_district_heating,GUELPA2021119440}. 

\begin{figure}[H]
\centering
    \begin{subfigure}[b]{0.45\textwidth}
        \centering
        \includegraphics[width=\linewidth , height=3.7cm]{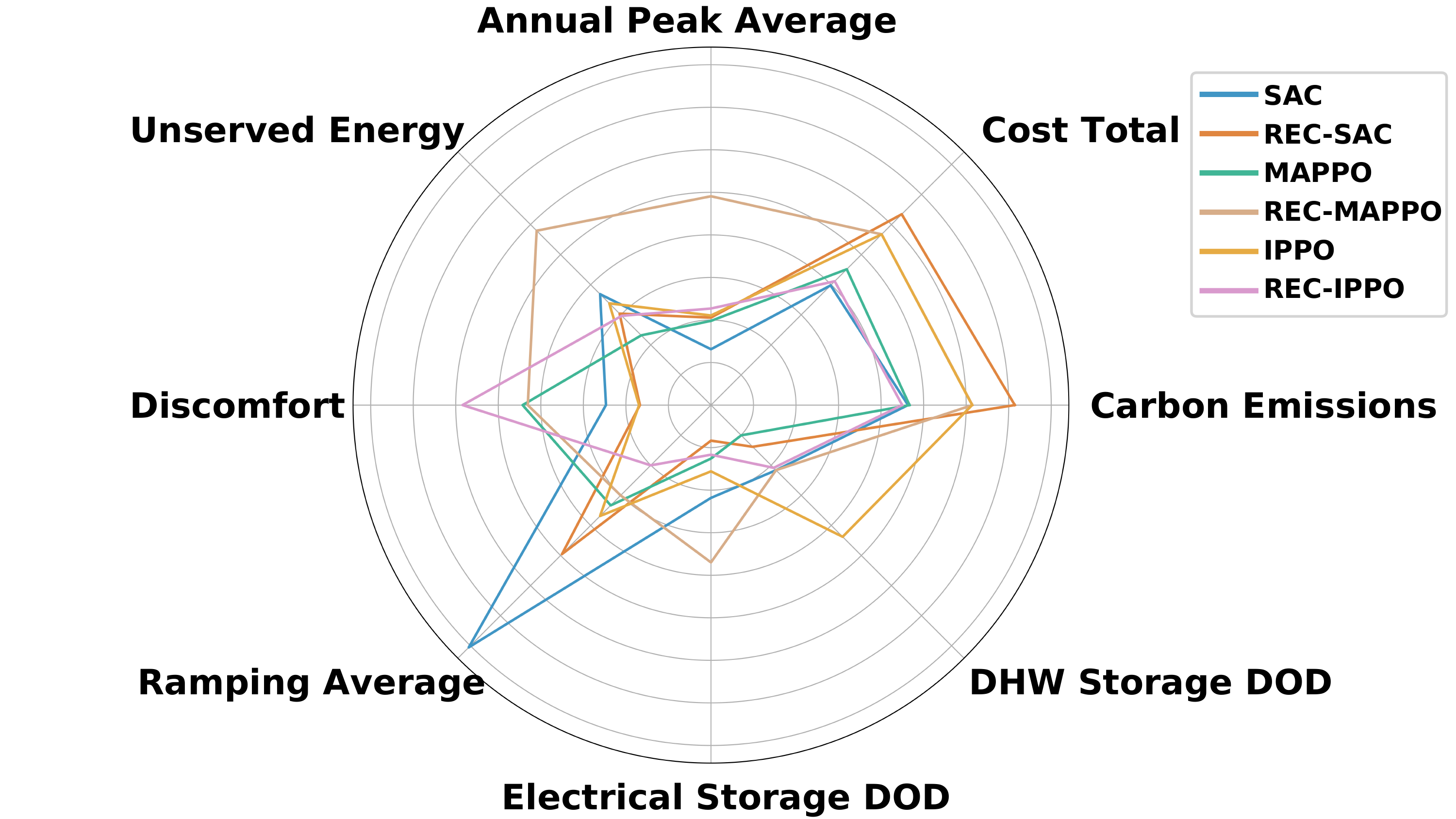}
        \caption{a) Multi-KPI performance comparison.}
        \label{fig:spider_graph}
    \end{subfigure}
    \begin{subfigure}[b]{0.45\textwidth}
        \centering
        \includegraphics[width=\linewidth]{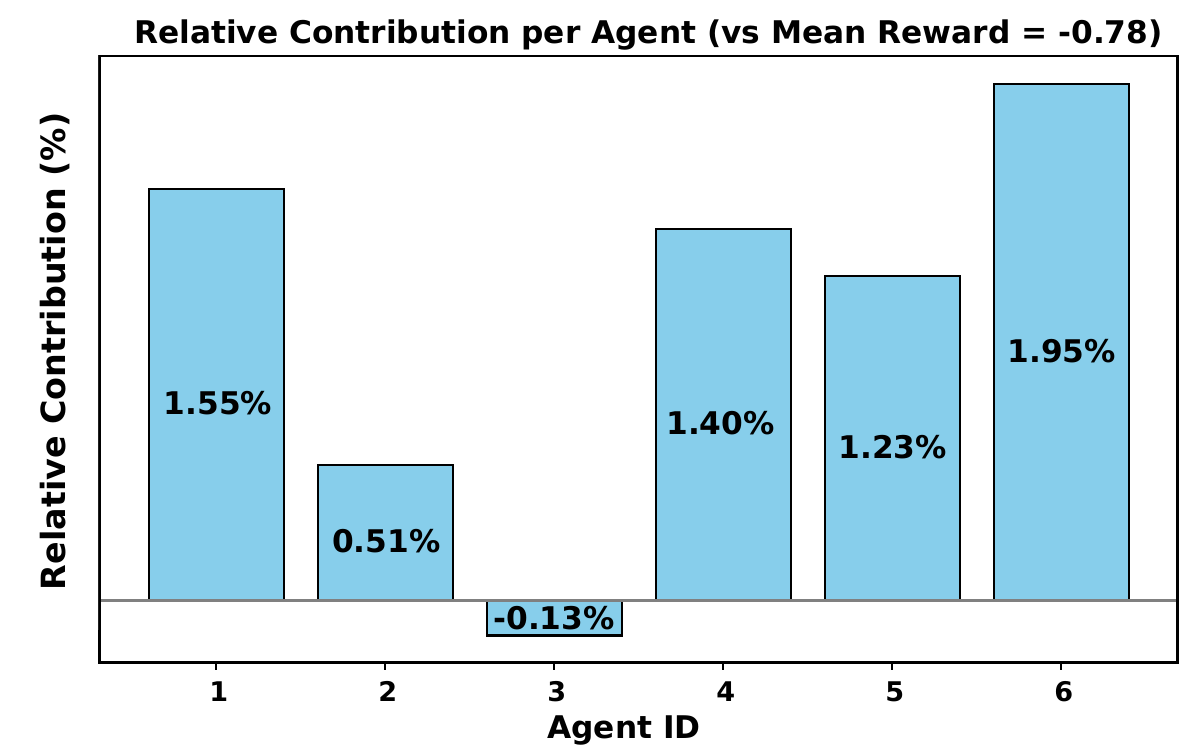}
        \caption{b) Example of per-agent contribution analysis.}
        \label{fig:relative_contribution}
    \end{subfigure}
    \hfill
    \begin{subfigure}[t]{0.45\textwidth}
        \centering
        \includegraphics[width=\linewidth]{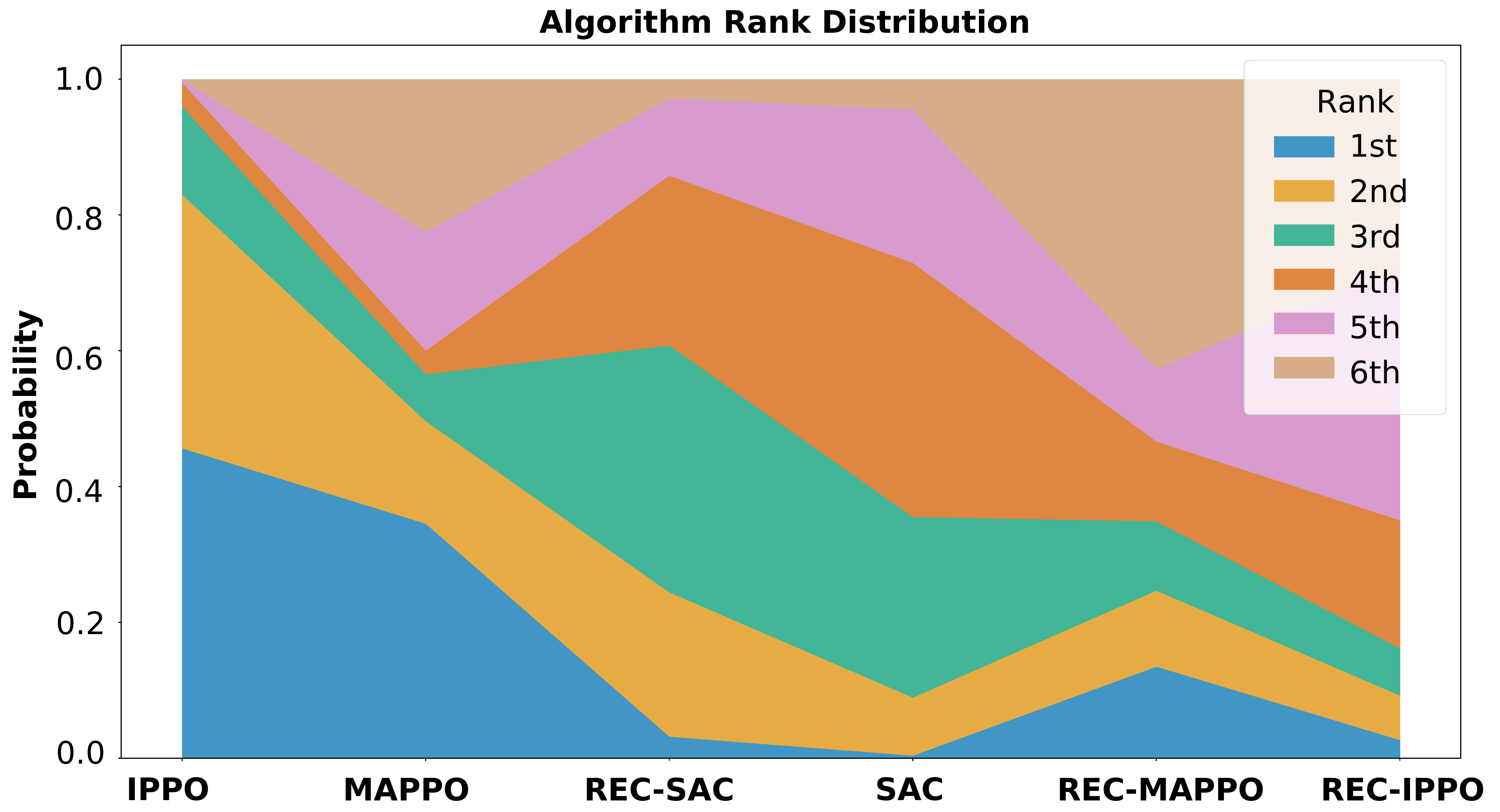}
        \caption{c) Probabilistic ranking across trials.}
        \label{fig:stackplot}
    \end{subfigure}
    \begin{subfigure}[t]{0.45\textwidth}
        \centering
        \includegraphics[width=\linewidth]{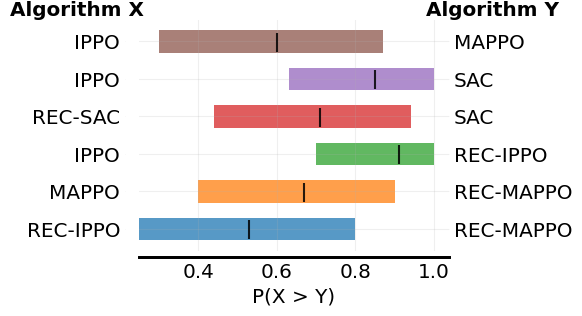}
        \caption{d) Pairwise probability of one algorithm outperforming another.}
        \label{fig:probofimp_score}
    \end{subfigure}

    \caption{
    \textbf{An illustrative overview of our comprehensive MARL evaluation framework.} (a) A multi-objective
comparison showing algorithm performance trade-offs across nine distinct KPIs.
(b) A sample analysis
quantifying granular metrics, such as the relative contribution of individual agents. 
(c) A probabilistic rank
distribution from our statistical robustness analysis, summarizing algorithm performance over multiple trial.
(d) Pairwise improvement probabilities indicating how likely one algorithm is to outperform another.}
\label{fig:introduction_plot}
\end{figure}

Transitioning to automated, intelligent control systems that can handle consumption variability, inter-dependencies among entities, and the optimization of conflicting objectives is critical. Given these requirements, Multi-Agent Reinforcement Learning (MARL) emerges as a promising framework for enabling intelligent, coordinated energy management at scale \citep{10.1145/3408308.3427604, 10.1145/3427773.3427870, 10.1145/3427773.3427869, pmlr-v220-nweye23a}. MARL extends the powerful framework of Reinforcement Learning to scenarios involving multiple interacting agents, enabling intelligent, coordinated decision making within dynamic environments. This makes it particularly effective for complex real world challenges by addressing scalability, non stationarity, and decentralized control, making it a highly relevant track to follow for urban energy systems. Beyond energy, MARL's capability for coordinated control is reflected in diverse real life applications: it is pivotal in autonomous driving for safe traffic flow \citep{zhang2024multiagentreinforcementlearningautonomous}, enables robotic swarms to collaborate on complex tasks like search and rescue \citet{BLAIS2023226}, dynamically allocates resources in telecommunications to optimize network performance \citep{Hady2025}, and develops sophisticated trading strategies with interacting algorithms in financial markets \citep{HUANG2024121502}.

Despite its potential, successfully applying MARL to real world energy management scenarios requires a thorough understanding of its strengths and limitations under varying conditions and system constraints. Addressing this gap, we present a comprehensive empirical comparison of six MARL algorithms to evaluate their effectiveness in urban energy management tasks. Our study focuses on two prominent paradigms:  Decentralized Training with
Decentralized Execution (DTDE), which treats each agent as an isolated entity without accounting for multi agent interactions \citep{10.5555/3091529.3091572} and Centralized Training with Decentralized
Execution (CTDE), which leverages centralized information during training while enabling decentralized decision making during operation. Furthermore, both feedforward and recurrent neural network architectures are employed to model agent observations, allowing us to assess the impact of temporal dependencies on algorithm performance.

The experiments are conducted within CityLearn \citep{citylearn, citylearn2}, an open source  environment designed to streamline and standardize the implementation, testing, and comparative evaluation of various control algorithms for building energy coordination and demand response in urban settings. The platform has been instrumental in fostering solutions to this problem, notably through the CityLearn Challenge \footnote{CityLearn Challenge, \url{https://www.citylearn.net/citylearn_challenge/index.html}} \citep{10.1145/3408308.3431122}, which invites participants to develop single agent, multi agent RL or model predictive control (MPC) policies. Past challenges have showcased diverse successful approaches: in the 2020 CityLearn Challenge, a centralized SAC algorithm secured second place \citep{10.1145/3427773.3427869}, while the 2022 challenge winners employed Multi Agent MAPPO for cooperative decision making. The fourth place team in the same 2022 challenge demonstrated the effectiveness of blending handcrafted policies with MARL to enhance interpretability and generalization \citep{pmlr-v220-nweye23a}.

This study seeks to expand on this growing body of work by providing a comprehensive benchmark of six MARL algorithms within the CityLearn environment. We systematically compare different training paradigms, specifically DTDE and CTDE approaches.

To offer a robust analysis, we focus on training algorithms representing both on-policy and off-policy learning, selecting Proximal Policy Optimization (PPO) as a key on-policy algorithm and Soft Actor-Critic (SAC) as a representative off-policy method. For each of these algorithms, we explore different training schemes to thoroughly investigate what coordination paradigm yields the most effective results (comparing DTDE vs. CTDE) and whether learning temporal dependencies can significantly enhance performance. We find that incorporating temporal dependencies is a key factor for improving performance on critical time-based metrics. Furthermore, we reveal a fundamental trade-off: decentralized learners provide superior stability and lower variability, whereas centralized variants, while less stable, can achieve significantly higher peak performance.

Our contributions are therefore summarized as follows.
\begin{itemize}
    \item \textbf{Rigorous Robustness Benchmarking:} We conduct extensive testing across multiple seeds with rigorous hyperparameter tuning, providing detailed benchmarks that systematically assess the performance variability and robustness of MARL algorithms within the CityLearn environment.
    \item \textbf{Comprehensive Evaluation Framework:} We introduce an evaluation methodology that extends standard KPIs (grid stability, cost, emissions) with novel metrics critical for real-world deployment, including battery lifetime, individual agent contribution, and worst-case performance.
    \item \textbf{In-depth Trade-off Analysis:} We present a critical, systematic comparison of the trade-offs and limitations of existing MARL algorithms, offering clear insights into their practical suitability for complex, multi-agent energy management tasks.
\end{itemize}

To visually encapsulate the scope of our analysis, \autoref{fig:introduction_plot} presents a snapshot of our comprehensive evaluation framework.



\section{Methodology}
\vspace{-0.25cm}The methodology detailed in this section outlines the algorithmic foundation, the unique challenges of the CityLearn environment, the KPIs used for evaluation, and the robust evaluation protocol used in all experiments.

\subsection{Algorithms}
All algorithms are implemented using the scalable Sebulba architecture \citep{podracer}, which supports efficient multi agent training and execution.  We evaluate both decentralized and centralized training paradigms across on-policy and off-policy methods, with each implemented in feedforward and recurrent variants to study the impact of temporal abstraction.

Specifically, under DTDE paradigm, we include Independent Soft Actor Critic (ISAC) \citep{haarnoja2018softactorcriticoffpolicymaximum} and Independent Proximal Policy Optimization (IPPO) \citep{yu2022surprisingeffectivenessppocooperative} . For CTDE, we evaluate Multi Agent Proximal Policy Optimization (MAPPO) \citep{yu2022surprisingeffectivenessppocooperative} where the global observation is formed by concatenating the observations of all agents.

To capture temporal dependencies, each method is implemented with a recurrent observation encoder using Gated Recurrent Units (GRUs), alongside a feedforward baseline. This allows for a direct comparison of the effect of temporal modeling, which is crucial in energy environments characterized with daily load cycles and delayed reward signals \citep{heess2015memorybasedcontrolrecurrentneural}.  All algorithms are trained under identical rollout conditions and a shared reward setting, where the same reward function applies across algorithms and all agents receive the same signal to encourage cooperative behavior.  Additional implementation and hyperparameter details are provided in the Appendix \hyperref[sec:AppendixB]{B}.

\subsection{CityLearn Environment}

\paragraph{Environment Overview.}
We use the CityLearn Challenge 2023 dataset \footnote{\href{https://dataverse.tdl.org/dataset.xhtml?persistentId=doi:10.18738/T8/SXFWTI}{Dataverse: DOI 10.18738/T8/SXFWTI}} \citep{T8/SXFWTI_2024}, which simulates a single family neighborhood composed of six buildings equipped with space cooling, domestic hot water (DHW)  heating, and energy storage. The original contest releases a three building “warm-up” set and later evaluates on a hidden three month trace, with performance reflected on a public leaderboard. In this study we bypass the phase split and train test on the complete three month, six building dataset to ensure consistent, full scale comparison (see Appendix \hyperref[sec:appendix_C]{C} for more details about the data).

\paragraph{Observation.}
Each agent receives a wide range of observations, including current and forecasted variables to anticipate future events. These include for example electricity prices, outdoor temperatures and energy demands. A complete list of input features is provided in Table \ref{tab:observations_description} of Appendix \hyperref[sec:appendix_C]{C}.
\paragraph{Actions.}
Each agent controls three continuous actions, normalized to the [0, 1] range:
\begin{itemize}
    \item \textbf{DHW storage.} 
    Controls the charge and discharge level of the hot water battery. Values below 0.5 correspond to discharging, and values above 0.5 to charging. A value of 0.2 discharges the battery at 60\% of its capacity.
    \item \textbf{Electrical storage.}
     Manages charging of the battery used for general electrical consumption (non shiftable load usage). The interpretation of action values is the same as for DHW storage.
    \item \textbf{Cooling device.} Controls the usage of the building's cooling system. An action value of 0.25 corresponds to using the device at 25\% capacity.
\end{itemize}
\paragraph{Reward Structuring.}
At each timestep, every agent receives a scalar reward that combines four key objectives related to comfort, efficiency, and sustainability. The reward for agent $i$ at time $t$ is defined as:
\begin{equation}
    r^i(t) = -(\alpha d(t)+\beta e(t)+\gamma r(t)+\lambda s(t)) 
\end{equation}
Here, $d(t)$ represents thermal discomfort, computed as the deviation between actual and desired indoor temperature. e(t) is the total electricity consumption at timestep t. $r(t)$ is a ramping penalty that captures the absolute change in electricity usage between two consecutive timesteps, encouraging smoother operation. s(t) is a solar penalty, defined as the difference between imported electricity and on site solar generation, penalizing under utilization of local renewable resources.

We set the weights \(\alpha\ = \beta = \gamma = \lambda = 0.1 \), giving the four objectives equal importance and keeping the overall reward scale simple and interpretable.

\subsection{Key Performance Indicators}
\textbf{}To judge how well each control strategy balances comfort, cost and sustainability, we track a core set of KPIs provided by CityLearn and augment them with three domain specific metrics.

\textbf{Built in KPIs.}
The CityLearn environment provides a set of standardized KPIs that quantify control performance across several real world objectives. These metrics are computed based on the environment’s state after agent actions are applied, and include for example carbon emissions, total electricity consumption, ramping effect, and occupant discomfort. All KPIs are defined such that lower values indicate better performance. A complete list of environment provided KPIs is included in \autoref{tab:kpi_description} of \hyperref[sec:appendix_C]{Appendix C}.

\textbf{Battery Depth of Discharge (DoD).}
This metric quantifies how aggressively agents use energy storage. Managing DoD is critical for battery health and lifespan, as deeper discharge cycles accelerate degradation. In our setting, fluctuating and intermittent renewable generation often prevents batteries from reaching a fully charged state, complicating traditional DoD computation. To address this, we apply a method based on rainflow cycle counting, originally developed in fatigue analysis, which breaks down complex charge discharge patterns into identifiable cycles. These cycles reveal discharge amplitudes and durations, enabling a clearer and more accurate proxy for long term battery wear. Higher DoD indicates deeper discharges and therefore faster aging.

\textbf{Agent Importance Score.}
The second metric, computed using Agent Importance \citep{efficientagentcontribution} , addresses the multi agent credit assignment, which is crucial in cooperative multi agent settings like CityLearn. In such cooperative settings, agents must coordinate to achieve shared objectives, but their influence on the outcome can vary significantly. This metric approximates each agent’s marginal impact by comparing the team’s reward with and without the actions of each agent, using a Shapley value \citep{Shapley1951}-inspired formulation adapted for computational efficiency. In doing so, it highlights whether an algorithm distributes responsibility effectively or relies heavily on a subset of agents. Crucially, this measure enables the detection of lazy agents that consistently contribute little or even negatively to team performance and provides insight into credit assignment and coordination quality across control methods.

\subsection{Evaluation Protocol}
\label{sec:section4_2}

To ensure fair comparison across algorithms, each algorithm underwent independent hyperparameter tuning within the environment. A total of 21 hyperparameter configurations were used in training, each using three random seeds. The final average score across 3 seeds was used to select the best performing configuration. The optimal combination was then fixed and used for the benchmarking phase to ensure fair comparison across algorithms. Complete search grids and the selected values for each method appear in  \hyperref[sec:AppendixB]{Appendix B}.

Following this model selection, we evaluate all algorithms using the CityLearn 2023 dataset, following a comprehensive protocol that captures both performance and robustness. Core metrics include the official average score from the CityLearn leaderboard, calculated as a weighted sum of normalized KPIs emphasizing occupant comfort and grid impact.

To assess statistical reliability, we report absolute metrics \citep{colas2018geppgdecouplingexplorationexploitation}, which evaluate performance using the best-performing model parameters over a substantially larger number of episodes—typically ten times more than standard assessments. This approach provides a more statistically reliable measure of performance, emphasizing consistency under optimal conditions while also revealing potential deviations and worst-case outcomes, thereby offering deeper insight into the algorithm’s robustness, and aggregate performance using the Interquartile Mean (IQM) and Conditional Value at Risk (CVaR) \citep{statprecipice, reliabilitymetrics}. IQM captures central tendency while reducing sensitivity to outliers, whereas CVaR quantifies performance in adverse conditions which is critical in high stakes environments like energy systems. Finally, we estimate 95\% confidence intervals using stratified bootstrap resampling, and complement point estimates with the probability of improvement metric and rank distribution plots for robust pairwise comparisons and ranking stability \citep{statprecipice}. We also include sample efficiency curves to visualize learning speed and stability over time.

\section{Results and Analysis}
In this section we compile the results across the six algorithms on citylearn environment. The algorithms are evaluated on the CityLearn 2023 dataset using the protocol described in Section \hyperref[sec:section4_2]{2.4}. Results are presented using aggregate and average scores, targeted KPI breakdowns, and additional diagnostic metrics. All plots are custom generated or adapted using open source benchmarking tools provided by \citet{emarl}, ensuring consistency in evaluation and reproducibility.

\subsection{Learning Efficiency and Convergence}

The average score aggregates several environment level KPIs, where lower values indicate better overall performance. SAC demonstrates faster initial improvement than other algorithms but plateaus early potentially linked to its entropy driven exploration. In contrast, feedforward IPPO converges more gradually yet ultimately achieves the lowest average score, reflecting superior long term average performance. Recurrent and feedforward MAPPO variants exhibit greater variability across seeds, as evidenced by wide standard deviation envelope, likely due to the increased input size of its centralized critic and sensitivity to joint observations. This is further exacerbated in the recurrent setting, where temporal dependencies introduce further instability.
These trends reveal the trade off between early sample efficiency (favored by SAC) and stable long term performance (achieved by IPPO).

\begin{figure}
    \centering
    \includegraphics[width=0.7\linewidth, height = 4.5cm]{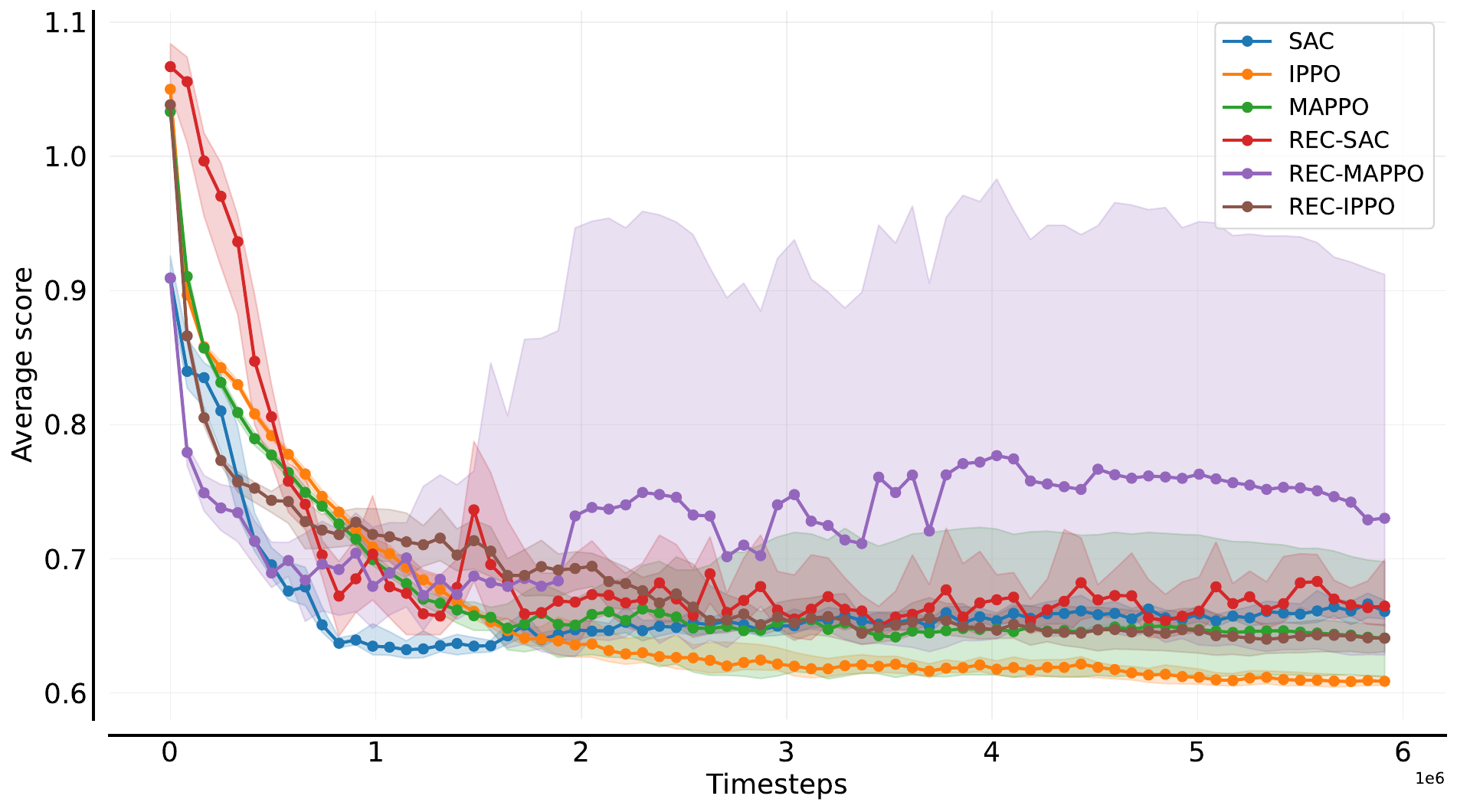}
    \caption{Sample efficiency curves for algorithms on the average score metric, showing SAC’s strong early gains but weaker plateaued performance, while IPPO ultimately achieves the lowest average score and best overall results.}
    \label{fig:Sampleeff}
\end{figure}

\begin{figure}[H]
    \centering
    \begin{subfigure}[b]{0.48\textwidth}
        \centering
        \includegraphics[width=0.8\linewidth, height=2.5cm]{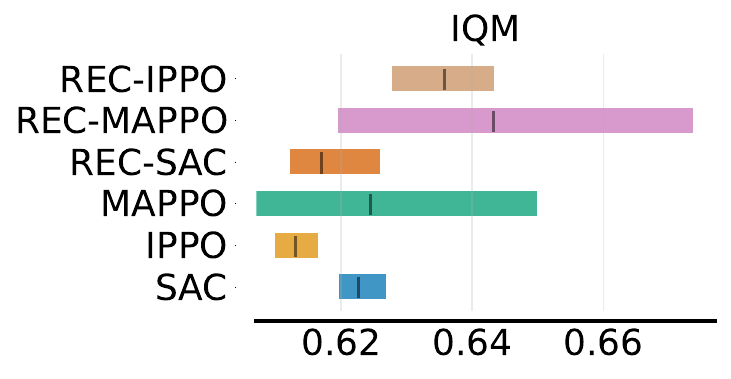}
        \caption{a) Aggregate Score (IQM)}
        \label{fig:aggscore}
    \end{subfigure}
    \hfill 
    \begin{subfigure}[b]{0.48\textwidth}
        \centering
        \includegraphics[width=0.8\linewidth, height=2.5cm]{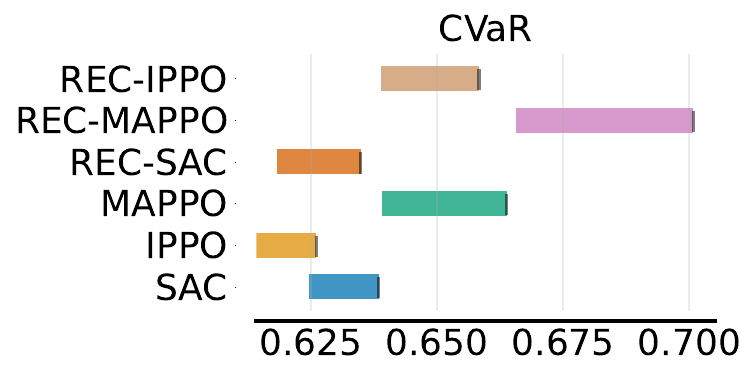}
        \caption{b) CVaR of Aggregate Score}
        \label{fig:cvaraggscore}
    \end{subfigure}

    \caption{
        \textbf{Aggregate Performance Analysis of Control Algorithms.} 
        This figure presents a consolidated view of algorithm performance across all evaluation scenarios. 
        \textbf{(a)} The IQM of the aggregate score, where lower is better. IPPO achieves the best average performance. 
        \textbf{(b)} The CVaR of the score highlights worst-case outcomes, again showing IPPO's superiority and indicating robust performance. The wide confidence intervals for MAPPO in both (a) and (b) reveal its high sensitivity to random seeds.
    }
    \label{fig:aggregate_performance}
\end{figure}
\subsection{Aggregate Performance} 
\autoref{fig:aggscore} and \autoref{fig:cvaraggscore} illustrate IPPO's dominant aggregate performance. IPPO consistently achieves the lowest IQM confirming its robust and superior average-case performance across seeds and scenarios. In contrast, MAPPO displays high seed-to-seed variability, evidenced by visibly wider confidence intervals than those of the independent learners. We attribute this instability to its high-dimensional, centralized critic, suggesting it could benefit from improved observation design (as discussed in Appendix \hyperref[appendix_E]{E}. Crucially, IPPO excels in both average-case and worst-case performance, underscoring its comprehensive robustness.
\FloatBarrier
\subsection{Targeted Environment KPI Level Comparison}
\begin{figure}
    \centering
    \begin{subfigure}[b]{0.3\textwidth}
        \centering
        \includegraphics[width=\linewidth]{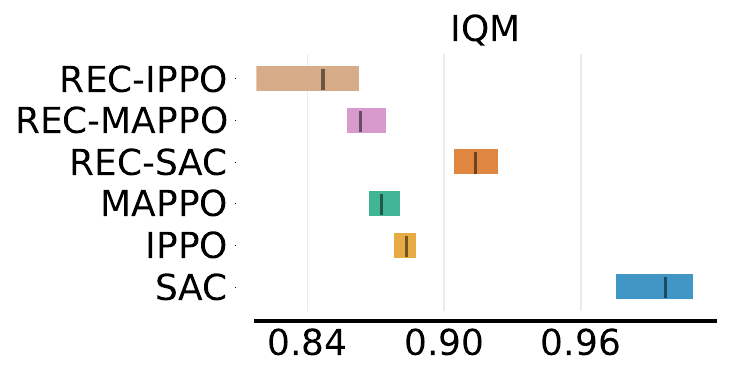}
        \caption{a) Ramping (IQM)}
        \label{fig:ramping}
    \end{subfigure}
    \hfill
    \begin{subfigure}[b]{0.3\textwidth}
        \centering
        \includegraphics[width=\linewidth]{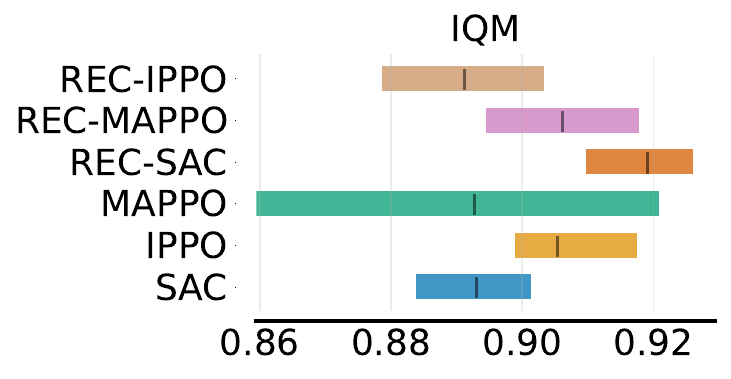}
        \caption{b) Carbon Emissions (IQM)}
        \label{fig:carbon}
    \end{subfigure}
    \hfill
    \begin{subfigure}[b]{0.3\textwidth}
        \centering
        \includegraphics[width=\linewidth]{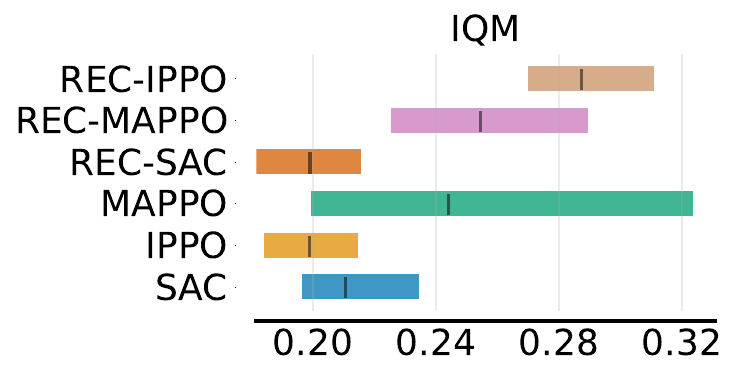}
        \caption{c) Discomfort Proportion (IQM)}
        \label{fig:discomfort}
    \end{subfigure}
    
    \vspace{0.2cm} 
    
    \begin{subfigure}[b]{0.3\textwidth}
        \centering
        \includegraphics[width=\linewidth]{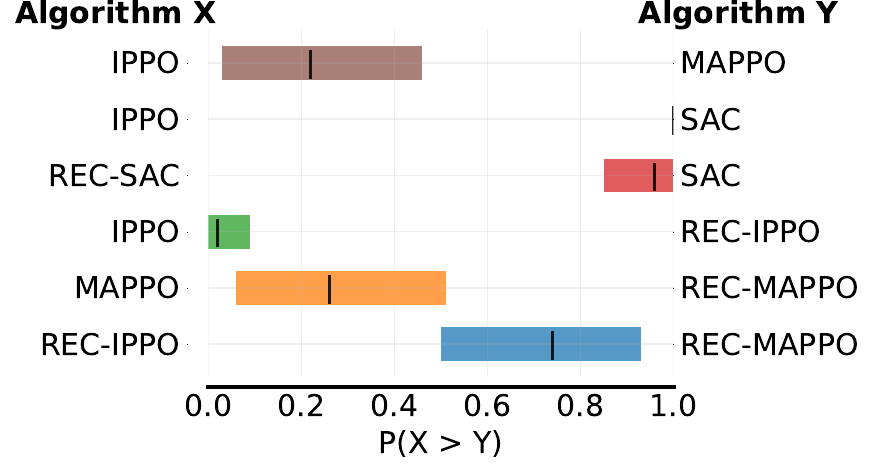}
        \caption{d) Ramping (Probability of Improvement)}
        \label{fig:impprob}
    \end{subfigure}
    \hfill
    \begin{subfigure}[b]{0.3\textwidth}
        \centering
        \includegraphics[width=\linewidth]{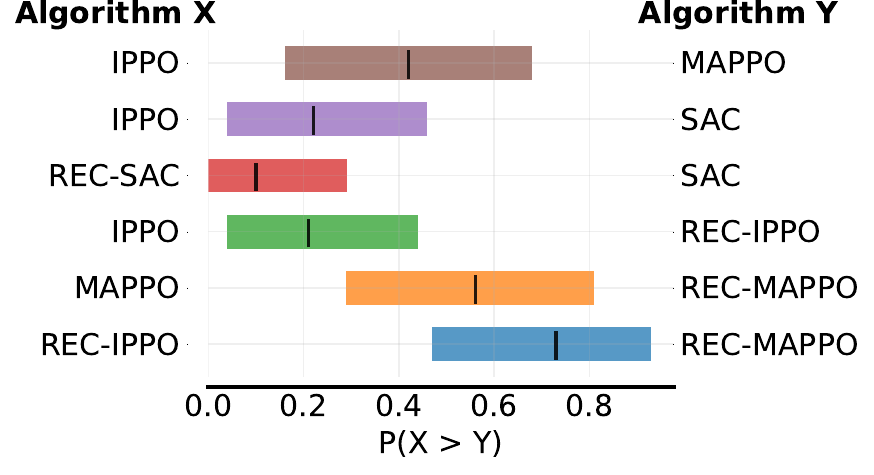}
        \caption{e) Carbon Emissions (Probability of Improvement)}
        \label{fig:carbon_imp}
    \end{subfigure}
    \hfill
    \begin{subfigure}[b]{0.3\textwidth}
        \centering
        \includegraphics[width=\linewidth]{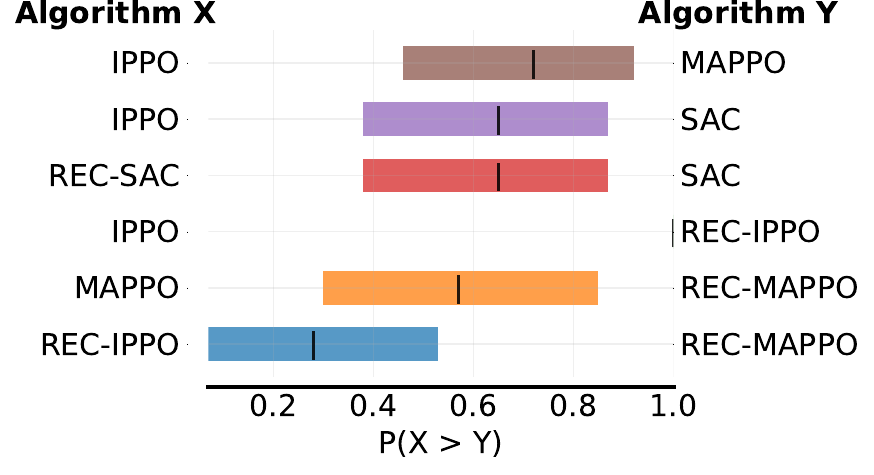}
        \caption{f) Discomfort Proportion (Probability of Improvement)}
        \label{fig:discomfort_imp}
    \end{subfigure}

    \caption{
    \textbf{Comparison of Key Performance Indicators (KPIs) Across Algorithms.}
    Performance on three environmental metrics, where lower IQM scores (a-c) are better.
    \textbf{(a) Ramping (IQM):} Recurrent models significantly outperform their feedforward counterparts on this metric.
    \textbf{(b) Carbon Emissions (IQM):} All algorithms perform similarly, with no clear, consistent advantage from temporal dependency.
    \textbf{(c) Discomfort Proportion (IQM):} Recurrent-SAC and IPPO achieve the strongest performance, highlighting decentralized variants' potential on this metric.
    \textbf{(d–f) Probability of Improvement:} These plots mirror the trends in (a–c), showing that recurrent models outperform feedforward variants on ramping (d), but offers no clear advantage on carbon emissions (e), and on discomfort reduction (f). Notably, beyond Recurrent-IPPO, the decentralized variants also exhibit a high probability of outperforming other approaches on the discomfort metric.
}
\label{fig:kpi_comparison}
\end{figure}

This subsection focuses on three key environment level KPIs provided by CityLearn such as ramping, carbon emissions, and discomfort proportion, as primary indicators of control quality. Results are based on \autoref{fig:carbon} to \autoref{fig:impprob}, which highlight these metrics across algorithms. Additional KPI results, including those for energy cost, load factor, and unserved energy, are provided in Appendix \hyperref[appendix_C]{C}.

\textbf{Ramping.} \autoref{fig:ramping} shows that temporal dependency leads to improved performance on the ramping KPI compared to their feedforward counterparts. This effect is most pronounced in Recurrent IPPO, which outperforms all other algorithms on this metric. This distinction is further highlighted by the probability of improvement graph shown in \autoref{fig:impprob}, which reveals that recurrent systems have a much higher likelihood of outperforming their feedforward counterparts for this metric. The improvement highlights the benefit of temporal memory: it enables agents to better model sequential dependencies and anticipate future demand, leading to smoother transitions in power consumption and reduced fluctuations between consecutive time steps.

This suggests that temporal dependency can be beneficial when memory is crucial to account for previous electricity usage. Furthermore, it appears to have a more pronounced positive effect on independent learners compared to centralized ones. This is particularly evident in the ramping KPI, where Recurrent IPPO shows a more substantial improvement over IPPO, and Recurrent SAC significantly enhances SAC's performance.

\vspace{0.1cm}

\textbf{Carbon emissions.}

Based on \autoref{fig:carbon}, all algorithms generally perform within a similar range for carbon emissions, with differences being relatively small. MAPPO occasionally achieves lower emissions than the others, while SAC shows more consistent results with a smaller confidence interval. The occasional superior performance of MAPPO may indicate that approaches which better handle global observations—beyond simple concatenation of agents observations—could further improve results. Overall, recurrent architectures do not provide a clear advantage for this metric, with models such as rec-SAC and rec-MAPPO sometimes underperforming their feedforward counterparts, suggesting that temporal dependency plays a limited role in reducing carbon emissions in these settings. This is further supported by the probability-of-improvement results, which show no consistent likelihood of recurrent models outperforming feedforward ones on this KPI.

\vspace{0.1cm}
\textbf{Discomfort Proportion.}

Recurrent-SAC and IPPO perform comparably on the discomfort KPI, as indicated by overlapping confidence intervals in \autoref{fig:discomfort}. More broadly, temporal dependency has a mixed impact on this metric: Recurrent SAC shows only a slight improvement over SAC, whereas the other recurrent algorithms perform worse. This suggests that discomfort minimization, which is driven by short-term deviations in thermal conditions, is primarily addressed through rapid, local adjustments rather than temporal modeling. Unlike ramping, which benefits from anticipating future dynamics, discomfort requires fast, reactive control. These findings reinforce the earlier trade-off: temporal dependency can improve performance on tasks with strong temporal structure, such as ramping, but offers minimal benefit for objectives dominated by immediate, localized responses. The probability-of-improvement plots align with this pattern, showing that decentralized variants—including IPPO and Recurrent-SAC—have a consistently higher likelihood of outperforming other approaches on this metric.

While this alignment with temporal control theory is intuitive, these results are critical for establishing the diminishing returns of model complexity. By quantifying the performance margins, we demonstrate that the computational overhead of recurrent architectures provides no functional advantage for reactive objectives like discomfort minimization, but could be justified for objectives with temporal control.

\subsection{Battery Depth of Discharge}
\begin{figure}
    \centering
    \begin{subfigure}[b]{0.48\textwidth}
        \centering
        \includegraphics[width=\linewidth, height=3cm]{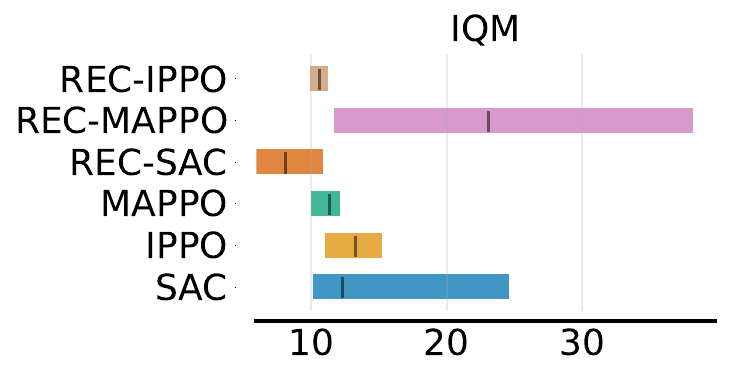}
        \caption{a) Electrical Storage - Depth of Discharge (DoD)}
        \label{fig:elecbatterydod}
    \end{subfigure}
    \hfill
    \begin{subfigure}[b]{0.48\textwidth}
        \centering
        \includegraphics[width=\linewidth, height=3cm]{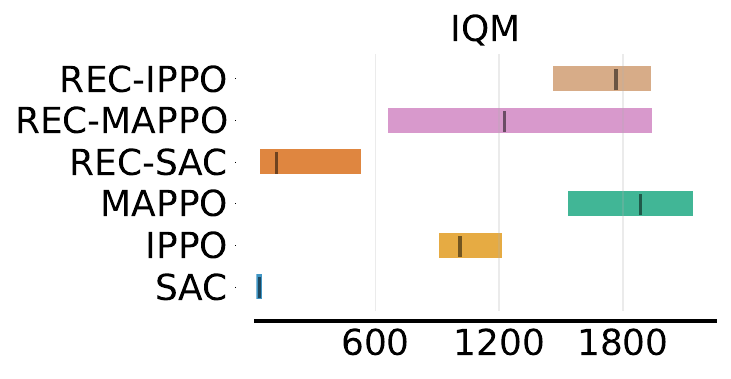}
        \caption{b) Electrical Storage - Discharge Duration}
        \label{fig:elecbatterydur}
    \end{subfigure}

    \begin{subfigure}[b]{0.48\textwidth}
        \centering
        \includegraphics[width=\linewidth, height=3cm]{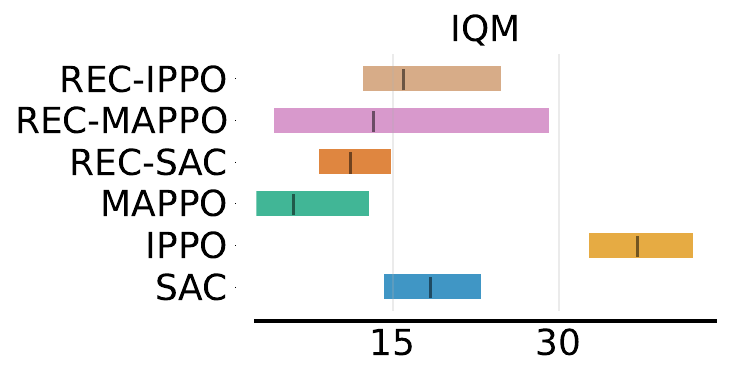}
        \caption{c) Hot Water Storage - Depth of Discharge (DoD)}
        \label{fig:dhwbatterydod}
    \end{subfigure}
    \hfill
    \begin{subfigure}[b]{0.48\textwidth}
        \centering
        \includegraphics[width=\linewidth, height=3cm]{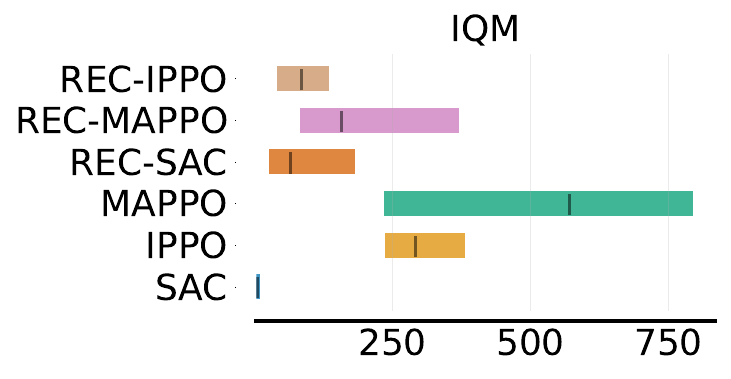}
        \caption{d) Hot Water Storage - Discharge Duration}
        \label{fig:dhwbatterydur}
    \end{subfigure}

    \caption{
        \textbf{Battery Usage Patterns for Electrical and Hot Water Storage.}
        This figure compares the battery management strategies learned by the algorithms, focusing on Depth of Discharge (DoD) and average discharge duration. Lower DoD IQM scores indicate less strain on the batteries while higher duration IQM scores indicate longer discharges which is more beneficial.
        \textbf{(a, c)} The DoD for both electrical and hot water storage is consistently lower for recurrent independent learners (Rec-IPPO and Rec-SAC), indicating shallower and less stressful discharge cycles.
        \textbf{(b, d)} Similarly, these recurrent agents achieve longer average discharge durations for electrical storage, reflecting more frequent and responsive battery usage.
    }
    \label{fig:battery_performance}
\end{figure}

\autoref{fig:elecbatterydod} through \autoref{fig:dhwbatterydur} show battery DoD and discharge duration across both electrical and hot water systems. Incorporating temporal dependency for independent learners, such as IPPO and SAC, consistently improves performance on both metrics, enabling discharge cycles that are not only longer but also shallower. This reflects more efficient use of storage capacity and suggests a potentially longer battery lifespan due to reduced strain per cycle.

These results align with earlier findings: temporal dependency is particularly beneficial for temporally structured control, enabling agents to better anticipate and adapt to dynamic conditions. The advantage is most pronounced in independent learners and Recurrent IPPO, in particular, mirrors its superior ramping performance with more deliberate and sustained battery usage.

In contrast, MAPPO shows less improvement from temporal dependency because its centralized critic evaluates the system at a global level rather than focusing on each agent’s local battery state. This reduces individual agents’ autonomy in managing their own storage, limiting their ability to fully exploit temporal patterns for optimizing local discharge behavior.

Importantly, DoD is not explicitly included in the reward function. The observed improvements in battery behavior likely arise as a side effect of optimizing related objectives such as ramping minimization and the solar penalty, which naturally incentivize more efficient storage use. In this sense, temporal dependency may allow agents to learn to manage batteries more effectively as a side effect of minimizing volatility and aligning with cleaner energy usage.

Overall, these results suggest that recurrence enables more temporally coherent storage strategies, contributing to operational efficiency and potentially promoting longer battery lifespan, even without direct optimization of degradation related metrics.

\subsection{Agent Importance}
\begin{figure}
    \centering
    \begin{subfigure}[b]{0.48\textwidth}
        \centering
        \includegraphics[width=0.92\linewidth]{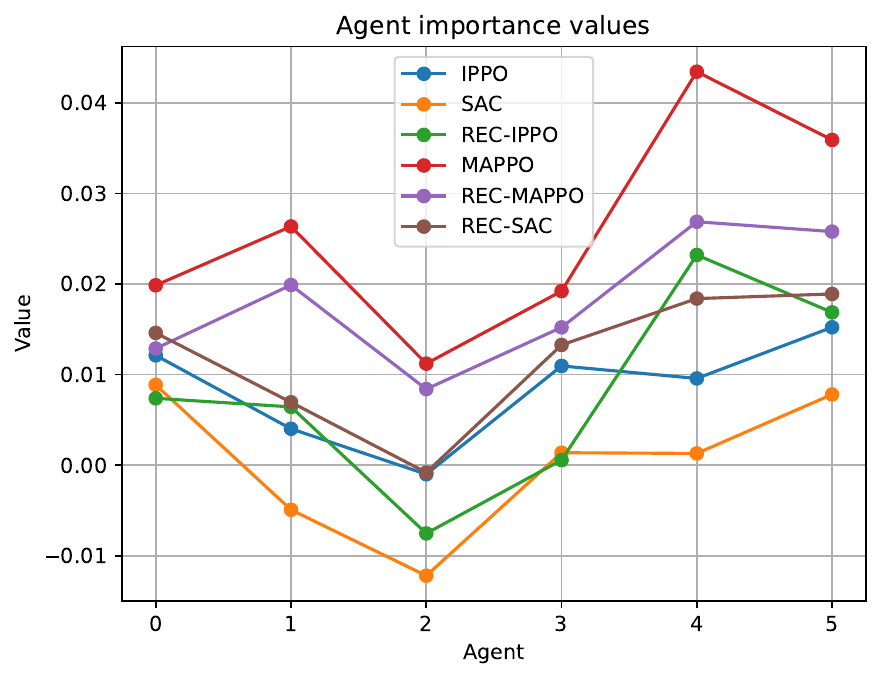}
        \caption{a) Agent Importance Scores}
        \label{fig:agentimp}
    \end{subfigure}
    \hfill 
    \begin{subfigure}[b]{0.48\textwidth}
        \centering
        \includegraphics[width=0.9\linewidth]{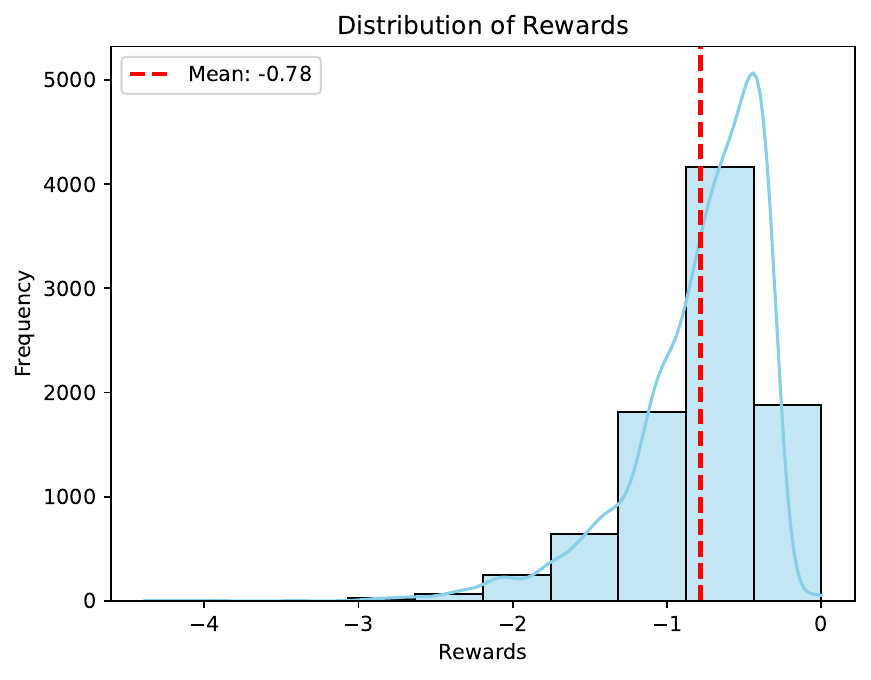}
        \caption{b) Distribution of Team Rewards}
        \label{fig:rewarddist}
    \end{subfigure}

    \caption{
    \textbf{Analysis of Individual Agent Contribution to Global Reward.}
    \textbf{(a)} Agent importance scores, which quantify individual influence on the total reward, are consistently near zero (typically ±0.05).
    \textbf{(b)} Distribution of the total team reward per time step, shown for scale.
    The comparison reveals that individual agent contributions (a) are negligible in magnitude relative to the overall team rewards (b).
}
    \label{fig:agent_contribution}
\end{figure}

\autoref{fig:agentimp} and \autoref{fig:rewarddist} show the agent importance scores and the distribution of team reward across time steps, respectively. Agent importance scores remain low (typically within ±0.05), and the broader scale of per step rewards indicates that such variations have negligible impact on overall performance. Together, these results suggest that no single agent plays a critical role in achieving the global reward. Removing any individual agent leads to only marginal performance degradation, implying that agents operate largely independently and that overall system behavior does not hinge on tight inter agent coordination, even in algorithms trained with centralized critics like MAPPO. Furthermore, we observe no evidence of “lazy agents” as contribution levels are generally balanced, and all agents participate meaningfully in achieving the shared objective. This points to effective credit distribution by the algorithms, especially those based on independent learning paradigms. This inherent decentralized robustness has significant real world advantages: in practical smart grid deployments agents may go offline, face communication issues, or operate under partial observability. The ability of the system to maintain high performance despite such disruptions makes these algorithms particularly well suited for large scale, fault tolerant applications in dynamic environments.
\clearpage
\section{Related Work}
\vspace{-0.25cm}
Reinforcement Learning (RL) has gained popularity in energy management due to its ability to optimize real time energy consumption and scale across DERs, demonstrating economic and environmental benefits \citep{en16145326, 9727169}. However, reproducibility remains a significant challenge in RL research due to various factors. These include variability in algorithm performance caused by inherent stochasticity, insufficient reporting of hyperparameters, and the absence of standardized testing environments. Such issues complicate cross study comparisons and pose significant barriers to replicating results \citep{emarl, statprecipice}. In urban enery management, standardized frameworks such as CityLearn address some of these issues by providing a consistent platform for algorithm evaluation \citep{citylearn}.

As part of this effort, the CityLearn Challenge \citep{10.1145/3408308.3431122} has become a go to benchmark for evaluating RL and hybrid control algorithms in urban energy scenarios. The challenge showcases a broad spectrum of solutions, from rule based controllers to pure RL and optimization based hybrids. For instance, the 2021 first place team from Virginia Tech employed a particularly elegant hybrid strategy: a linearized MPC problem served as the underlying policy, while an evolutionary algorithm tuned cost weights based on historical performance data. This approach achieved strong performance across multiple objectives such as peak load reduction, cost minimization, and emissions by blending interpretability with adaptability \citep{khattar2022winningcitylearnchallengeadaptive}.

However, managing multiple buildings with distinct profiles and operational constraints inherently constitutes a multi-agent problem. In this context, MARL emerges as a more natural and scalable alternative. CityLearn’s environment was designed to support this paradigm, allowing each building to be represented by an independent agent that can learn and act autonomously while coordinating with others. The capability of MARL was in particular evident in a later iterations of the CityLearn Challenge. In 2022, the winning team leveraged MAPPO to enable cooperative decision making among buildings. Meanwhile, another top performing team combined handcrafted rules with MARL to balance performance with transparency, showing that hybrid approaches could retain both control precision and interpretability \citep{pmlr-v220-nweye23a}. Such results highlight not only the flexibility of MARL but also the emerging design space between pure learning and domain informed heuristics.

Beyond the challenge setting, several independent works have further pushed the boundaries of MARL in energy control. MARLISA \citep{10.1145/3408308.3427604}, for example, employs a decentralized SAC framework with an iterative sequential action selection mechanism, coordinating agents while preserving decentralized execution. Similarly, MERLIN \citep{NWEYE2023121323} explores the effectiveness of training independent RL agents tailored to individual building profiles. This study found that personalized policies not only outperformed a baseline rule based controller but also led to improved district wide performance metrics which is a strong argument for decentralization done right.

However, this significant progress in developing novel MARL control architectures is currently undermined by a critical gap in rigorous, standardized benchmarking for complex domains like CityLearn. While prior work such as \citep{10.1145/3427773.3427870} has contributed foundational evaluations, there remains room to improve the consistency, reproducibility and statistical reliability of such evaluations mandated by modern MARL literature . To advance the field, benchmarking must address several points:  broader algorithmic coverage, increased evaluation runs to properly account for stochastic variability, and comprehensive reporting of hyperparameters and experimental settings as emphasized in recent literature \citep{emarl,statprecipice} .


\section{Conclusion}
This study benchmarks six MARL algorithms on the CityLearn environment, evaluating their performance across standard and custom KPIs relevant to energy management. The results highlight distinct trade offs: SAC excels in minimizing carbon emissions, while IPPO achieves strong and stable performance across both IQM and worst case metrics. Temporal dependency improves performance on temporally structured tasks like ramping and battery usage, but often underperform on short horizon objectives such as discomfort.
MAPPO variants show higher variability, reflecting the complexity of centralized coordination. Agent Importance analysis reveals that contributions are well distributed across agents, with no evidence of “lazy” agent behavior and with minimal drop in reward, even when individual agents or their access to key resources is removed, which supports the robustness and deployability of decentralized approaches. This reveals as well that overall system behavior does not depend on tight inter agent coordination, even under centralized critics like MAPPO. 
Furthermore, temporal dependency improves battery DoD and discharge duration, enabling more effective and sustained storage use. These gains arise indirectly from optimizing related objectives such as ramping minimization and solar penalty usage utilization, highlighting the benefit of temporal abstraction for more battery conscious control strategies.
Overall, the results underscore the importance of evaluating MARL algorithms along multiple dimensions such as average performance, risk sensitivity, coordination, and long term planning. 

\textbf{Future work.} Future work should explore reward shaping, observation design, and newer architectures such as value decomposition and attention-based models—including multi-agent transformers \citep{wen2022multiagentreinforcementlearningsequence} and Sable \citep{mahjoub2025sableperformantefficientscalable}—for further gains in robustness and scalability.

\bibliographystyle{IEEEtranN}
\bibliography{bib_file}

\newpage
\appendix
\section*{Appendix A}
This appendix provides brief descriptions of the algorithms used in our study.
\label{sec:AppendixA}
\subsection*{Algorithms}
All algorithms are implemented following the scalable Sebulba architecture \citep{podracer}, which facilitates efficient training and execution in large multi agent environments. A comprehensive table of used algorithms is provided in \autoref{tab:marl_taxonomy}
\begin{table}[h!]
    \centering
    \caption{Overview of MARL Algorithms Considered in This Study} 
    \label{tab:marl_taxonomy}
    \begin{tabular}{>{\raggedright\arraybackslash}p{3.5cm} 
                    >{\raggedright\arraybackslash}p{3.5cm} 
                    >{\raggedright\arraybackslash}p{3.5cm} 
                    >{\raggedright\arraybackslash}p{3.5cm}
    }
        \rowcolor{gray!20}
        \textbf{Policy Type} & \textbf{Coordination} & \textbf{Network Type} & \textbf{Algorithm} \\

        On-policy & Centralized & Feed Forward & MAPPO \\
        \rowcolor{gray!10}
        On-policy & Centralized & Recurrent & Rec-MAPPO \\
        On-policy & Decentralized & Feed Forward & Independent PPO (IPPO) \\
        \rowcolor{gray!10}
        On-policy & Decentralized & Recurrent & Rec IPPO \\
        Off-policy & Decentralized & Feed Forward & SAC \\
        \rowcolor{gray!10}
        Off-policy & Decentralized & Recurrent & Rec SAC \\
    \end{tabular}
\end{table}

\subsection*{Decentralized Training Decentralized Execution (DTDE)}
It is known also as the category of Independent Learning (IL) \citep{10.5555/3091529.3091572} where each agent learns independently and treats the other agents as part of the environment.

\textbf{ISAC}: Independent Soft Actor Critic (ISAC) is a decentralized multi‑agent extension of Soft Actor Critic (SAC) \citep{haarnoja2018softactorcriticoffpolicymaximum}. Each agent has its own actor and critic networks, enabling the agents to learn policies based on their local observations while optimizing a shared global reward. Its entropy regularized objective encourages diverse action selection, promoting both stable convergence and improved exploration in complex environments.

\textbf{IPPO}: Independent Proximal Policy Optimization (IPPO) \citet{yu2022surprisingeffectivenessppocooperative} adapts the PPO algorithm \citep{schulman2017proximalpolicyoptimizationalgorithms} for decentralized training. It uses a surrogate objective that clips policy updates to prevent drastic changes, ensuring stable learning. This approach allows for multiple update epochs on the same batch, enhancing sample efficiency and supporting individual agent adaptation in shared environments.

\subsection*{Centralized Training Decentralized Execution (CTDE)}
Unlike DTDE, CTDE lets agents share global information during training, typically through a joint critic, yet the deployed policies still act purely on their own local observations, so execution remains fully decentralised.

\textbf{MAPPO}: Multi Agent Proximal Policy Optimization (MAPPO) \citet{yu2022surprisingeffectivenessppocooperative}  extends IPPO with a centralized critic that estimates a joint value function from the aggregated observations of all agents, providing a richer training signal that captures inter agent dependencies. As in PPO, MAPPO supports multiple gradient steps per batch and leverages a clipped surrogate objective, resulting in improved stability and sample efficiency.

\subsection*{On policy and Off policy}

\textbf{On policy methods}: On policy learning methods updates the policy with trajectories generated by that same policy. This results in stable but less sample efficient learning, exemplified by IPPO and MAPPO.

\textbf{Off policy methods}: Off policy algorithms can learn from data generated by any policy, enabling experience reuse and thus achieving greater sample efficiency. However, this comes at the cost of increased algorithmic complexity due to challenges in aligning the behavior and target policies \citep{haarnoja2018softactorcriticoffpolicymaximum}.

\subsection*{Feed forward and Recurrent Observation Encoders}
To assess the impact of temporal abstraction, each algorithm is implemented in two variants: a feedforward version and a recurrent version. 

\textbf{Feedforward Observation Encoder}:
Feedforward models use fully connected layers to map observations to actions and value estimates, and are suitable for environments where decisions depend primarily on current observations.

\textbf{Recurrent Observation Encoder}:
 The primary change introduced in these models is the use of recurrent neural networks, rather than feedforward networks in both the actor and critic networks, enabling agents to capture temporal dependencies over time, which is beneficial for tasks requiring consideration of past information \citep{heess2015memorybasedcontrolrecurrentneural}.

\textbf{Recurrent IPPO}, \textbf{Recurrent ISAC}, and \textbf{Recurrent MAPPO} extend their feedforward counterparts by replacing the feedforward encoder layers with recurrent ones. This modification allows agents to condition their decisions on past information, which is particularly beneficial in environments with sequential patterns such as daily load or generation cycles.
\newpage
\appendix
\section*{Appendix B}
Below, we provide details about the hyperparameter tuning process and the selected hyperparameters. 
\label{sec:AppendixB}
\subsection*{Hyperparameter Tuning}

To ensure fair and effective comparisons across algorithms, hyperparameter sweeps were conducted. Each algorithm was evaluated over a predefined set of values, with the best performing configuration selected for final training. This process minimizes the risk of performance differences arising purely from suboptimal tuning, rather than algorithmic capability. The values swept over and the selected values are detailed in \autoref{tab:hyperparam_sweep1}-\autoref{tab:experimental_setup} It is important to note that not all parameters are applicable to all types of algorithm.

\begin{table}[ht!]
    \centering
    \caption{Hyperparameter Sweep for PPO systems} \label{tab:hyperparam_sweep1}
    \begin{tabular}{>{\raggedright\arraybackslash}p{6cm}>{\centering\arraybackslash}p{6cm}}
        \rowcolor{gray!20}
        \textbf{Hyperparameter} & \textbf{Values Swept} \\ 
        \textbf{Update Epochs ($\gamma$)} & [2, 4, 8] \\
        \rowcolor{gray!10}
        \textbf{Actor Learning rate ($\alpha$)} & [1e-4, 2.5e-4, 5e-4] \\
        \textbf{Critic Learning rate ($\alpha$)} & [1e-4, 2.5e-4, 5e-4] \\
        \rowcolor{gray!10}
        \textbf{Clipping value} & [0.2, 0.1, 0.05] \\
        \textbf{Entropy Coefficient} & [0.0, 1e-2, 1e-55] \\
        \rowcolor{gray!10}
        \textbf{Max grad norm} & [0.5, 5, 10] \\
        \textbf{Number of minibatches} & [2, 4, 8] \\
        \rowcolor{gray!10}
        \textbf{Actor Network Layers} & [(128, 128), (256,256)] \\
        \textbf{Critic Network Layers} & [(128, 128), (256,256)] \\
    \end{tabular}
\end{table}

\begin{table}[ht!]
    \centering
    \caption{Hyperparameter Sweep for SAC systems} \label{tab:hyperparam_sweep}
    \begin{tabular}{>{\raggedright\arraybackslash}p{6cm}>{\centering\arraybackslash}p{6cm}}
        \rowcolor{gray!20}
        \textbf{Hyperparameter} & \textbf{Values Swept} \\ 
        \textbf{Minibatch size} & [64, 128, 256, 512] \\
        \rowcolor{gray!10}
        \textbf{Learning rate ($\alpha$)} & [1e-4, 2.5e-4, 5e-4] \\
        \textbf{Replay Buffer size} & [1e6, 5e6] \\
        \rowcolor{gray!10}
        \textbf{Actor Parameter update frequency} & [1, 2, 4] \\
        \textbf{Learner Start} & [23040, 46080] \\
        \rowcolor{gray!10}
        \textbf{Actor Network Layers} & [(128, 128), (256,256)] \\
        \textbf{Critic Network Layers} & [(128, 128), (256,256)] \\
    \end{tabular}
\end{table}
\begin{table}[ht!]
    \centering
    \caption{Experimental Setup for Algorithms} \label{tab:experimental_setup}
    \begin{tabular}{>{\raggedright\arraybackslash}p{4cm}>{\centering}p{1.3cm}>{\centering\arraybackslash}p{1.3cm}>{\centering\arraybackslash}p{1.3cm}>
    {\centering\arraybackslash}p{1.3cm}>{\centering\arraybackslash}p{1.3cm}>
    {\centering\arraybackslash}p{1.3cm}}
        \rowcolor{gray!20}
        \textbf{Experimental setup} & \textbf{SAX} & \textbf{Rec SAX} & \textbf{IPPO} & \textbf{REC IPPO} & \textbf{MAPPO} & \textbf{REC MAPPO} \\ 
        
        \textbf{Hyperparameters} & & & & & &\\
        \rowcolor{gray!10}
        Discount factor &0.95 &0.95 & 0.99 &0.99 & 0.99 &0.99 \\
        Batch size &8192 &8192 & 8192&8192 & 8192&8192 \\
        \rowcolor{gray!10}
        Replay buffer size &1M &1M &-&- &-&-  \\
        Minimum replay buffer size before updating &23040 &46080 &-&-&-&- \\
        \rowcolor{gray!10}
        Target smoothing coefficient &0.001 &0.001 &-&-&-&- \\
        
        Value Network architecture &Feed Forward &Recurrent &Feed Forward&Recurrent &Feed Forward&Recurrent \\
        \rowcolor{gray!10}
        Value Network Layer size &[256,256] &[256,256] &[256,256] &[256,256] &[256,256] &[256,256] \\
        
        Parameter sharing &True &True &True&True &True&True \\
        \rowcolor{gray!10}
        Optimiser (type, parameters) &Adam &Adam &Adam &Adam &Adam &Adam \\
        
        Learning rate &0.0025 & 0.0005 &0.0025 &0.0025  &0.0025 &0.0025 \\
        \rowcolor{gray!10}
        Seed range  &  \multicolumn{6}{>{\centering\arraybackslash}p{10cm}}
        {[1,100,432,700,1500,1800,4000,6200,7000,8000]} \\ 
        
        \vspace{0.05cm}
        \textbf{Computational resources} & \multicolumn{6}{>{\centering\arraybackslash}p{10cm}}{} \\
        \rowcolor{gray!10}
        Average Wall-clock time per algorithm & \multicolumn{6}{>{\centering\arraybackslash}p{10cm}}{4 hours} \\
        
        CPUs per experiment & \multicolumn{6}{>{\centering\arraybackslash}p{10cm}}{30} \\
        \rowcolor{gray!10}
        GPU per experiment & \multicolumn{6}{>{\centering\arraybackslash}p{10cm}}{1} \\
        
        RAM per experiment & \multicolumn{6}{>{\centering\arraybackslash}p{10cm}}{90GB} \\
        
        \rowcolor{gray!10}
        \textbf{Evaluation protocol} & \multicolumn{6}{>{\centering\arraybackslash}p{10cm}}{} \\ 
        
        Total training (timesteps) & \multicolumn{6}{>{\centering\arraybackslash}p{10cm}}{6M} \\
        \rowcolor{gray!10}
        Evaluation interval (timesteps) & \multicolumn{6}{>{\centering\arraybackslash}p{10cm}}{81920} \\
        
        Independent evaluation episodes & \multicolumn{6}{>{\centering\arraybackslash}p{10cm}}{64}  \\
        \rowcolor{gray!10}
        Absolute metric (evaluation episodes, aggregation method) & \multicolumn{6}{>{\centering\arraybackslash}p{10cm}}{640} \\
        
        Local aggregation method & \multicolumn{6}{>{\centering\arraybackslash}p{10cm}}{Mean} \\
        \rowcolor{gray!10}
        Global aggregation method & \multicolumn{6}{>{\centering\arraybackslash}p{10cm}}{Mean}  \\
        \vspace{0.05cm}
        \textbf{Environment settings} & \multicolumn{6}{>{\centering\arraybackslash}p{10cm}}{}  \\
        \rowcolor{gray!10}
        Environment name (version) & \multicolumn{6}{>{\centering\arraybackslash}p{10cm}}{Citylearn v2.2b} \\
    
    \end{tabular}
\end{table}
\newpage
\newpage
\clearpage
\appendix
\section*{Appendix C}
\label{appendix_C}
To clarify the experimental setup, this appendix provides a detailed description of the environment used throughout our study. While the main text offers a high-level overview, the following sections outline the specifics of the observation and KPIs found in the environment.
\subsection*{Environment}
\subsubsection*{Data}

\label{sec:appendix_C}
The dataset includes multiple files that help provide observations for the simulation, each serving a specific purpose within the environment:
\begin{itemize}
    \item schema.json: This is a JSON file that models the CityLearn environment, loaded at runtime to define how the simulation operates. It includes buildings properties such as the devices and observations they have access to.
    \item weather.csv: This file contains real meteorological data corresponding to the location of the buildings. It is used to supply weather-related observation values to the environment, impacting simulations such as heating, cooling, and energy generation.
    \item carbon\_intensity.csv: This file provides the actual carbon emissions data associated with the grid's energy mix, measured in kilograms of CO2 equivalent per kilowatt-hour (kgCO2e/kWh). This data is crucial for evaluating the environmental impact of different energy management strategies.
    \item pricing.csv: This file includes the Time-of-Use (TOU) electricity rates, expressed in dollars per kilowatt-hour (\$/kWh). These rates fluctuate based on the time of day and are used to simulate economic incentives for energy consumption and savings.
    \item Building\_1...n.csv: These files contain time series data for each building in the dataset. The data includes various temporal variables, energy end-use demands, solar generation outputs, and indoor environmental variables.

    \item Building\_1...n.pth: These files store the parameters of Long Short-Term Memory LSTM models used to simulate the temperature dynamics within each building. 
    
\end{itemize}
\subsubsection*{Observations}
The observations received by each agent are comprehensive and include both real-time data and future predictions based on the provided dataset and simulations through the buildings' LSTM models.

The complete list of observations is detailed in \autoref{tab:observations_description}, which includes variables such as current energy demands, temperatures, and electricity prices. Notably, some observations feature variations representing predicted future states. For example, electricity pricing predictions are provided at multiple intervals: 6 hours, 12 hours, and 24 hours ahead. These predictions are crucial for enabling agents to anticipate changes in energy costs and optimize their strategies accordingly.

\begin{table}[ht!]

\begin{tabular}{>{\raggedright\arraybackslash}m{4cm} >{\raggedright\arraybackslash}m{10cm}}
\toprule
\textbf{Observation} & \textbf{Description} \\
\midrule
day type & Indicates the type of day (e.g., weekday, weekend, holiday), impacting building occupancy and energy consumption patterns. \\
\addlinespace
hour & Represents the hour of the day, crucial for daily patterns in electricity consumption, solar generation, and pricing. \\
\addlinespace
outdoor dry bulb temperature and variations & Represents outdoor temperature, informing anticipatory actions for heating or cooling systems. \\
\addlinespace
diffuse solar irradiance and variations & Measures sunlight from the sky, excluding direct sunlight, impacting natural light and heat in a building. \\
\addlinespace
direct solar irradiance and variations & Represents sunlight directly from the sun's direction, affecting solar power generation and building cooling needs. \\
\addlinespace
carbon intensity & Indicates carbon dioxide emitted per unit of electricity consumed, guiding decisions to reduce a building's carbon footprint. \\
\addlinespace
indoor dry bulb temperature & Represents indoor temperature, crucial for maintaining occupant comfort. \\
\addlinespace
non shiftable load & Indicates electrical loads that can't be altered or shifted in time, such as essential services. \\
\addlinespace
solar generation & Amount of electricity generated from solar panels. \\
\addlinespace
dhw storage soc & Represents the state of charge of the DHW storage system, indicating hot water availability. \\
\addlinespace
electrical storage soc & Shows the state of charge of an electrical storage system, such as a battery. \\
\addlinespace
net electricity consumption & Represents the difference between electricity consumption and generation, indicating the building's net demand. \\
\addlinespace
electricity pricing and variations & Indicates current and forecasted electricity prices, guiding decisions on when to consume or store energy. \\
\addlinespace
cooling demand & Amount of cooling required to maintain indoor temperature and comfort levels. \\
\addlinespace
dhw demand & Represents the building's hot water demand. \\
\addlinespace
occupant count & Indicates the number of people in the building, affecting energy use and comfort requirements. \\
\addlinespace
indoor dry bulb temperature set point & The target or desired indoor temperature. \\
\addlinespace
power outage & Indicates if there's a power outage, triggering backup systems or altering energy consumption behaviors. \\
\addlinespace
\bottomrule
\end{tabular}
\caption{Description of the observations that the agent recieves.}

\label{tab:observations_description}
\end{table}

\subsubsection*{KPIs}

To evaluate the performance of energy management systems effectively, a set of KPIs has been established. These KPIs serve as crucial metrics for assessing the efficiency, sustainability, and comfort levels of energy consumption within the district. By systematically measuring these indicators, insights into how well various algorithms and strategies perform in real-world scenarios are gained, facilitating informed decisions to optimize energy management practices. The following table outlines each KPI along with its corresponding description, highlighting their importance in the context of energy management within the CityLearn environment.

\begin{table}
\centering
\begin{tabular}{>{\raggedright\arraybackslash}m{5cm} >{\raggedright\arraybackslash}m{9cm}}
\toprule
\textbf{KPI Name} & \textbf{Description} \\
\midrule
\textbf{Carbon Emission (G)} & Total carbon emissions resulting from imported electricity, quantifying the environmental impact of energy consumption. Lower values indicate more environmentally friendly patterns. \\
\addlinespace
\textbf{Discomfort (U)} & Proportion of times when a building's indoor temperature is outside the comfort zone. A higher value indicates more discomfort for occupants. \\
\addlinespace
\textbf{Ramping (R)} & Average change in electricity consumption from one time step to another. Lower values suggest smoother transitions and fewer spikes, indicating a more stable power consumption profile. \\
\addlinespace
\textbf{Daily one minus load factor average (L)} & Average measure of the ratio of daily average electricity consumption to daily peak consumption. Closer to 1 indicates more even consumption throughout the day, while closer to 0 suggests variability. \\
\addlinespace
\textbf{Daily peak average (\(P_d\))} & Average of the maximum electricity consumption recorded at any time step in a day. \\
\addlinespace
\textbf{Annual Peak Average (\(P_n\))} & Highest electricity consumption value recorded across all time steps in a year. \\
\addlinespace
\textbf{One minus thermal Resilience (M)} & Measure of discomfort during power outages. A higher value indicates less resilience in maintaining comfortable temperatures during outages. \\
\addlinespace
\textbf{Power Outage Normalized unserved Energy (S)} & Proportion of unmet energy demand during power outages, normalized against total energy requirement. Highlights energy deficits during interruptions. \\
\textbf{Zero Net Energy (N)} & The proportion of time the district achieves zero net energy consumption. Zero Net Energy means that the total amount of energy used by the district is equal to the amount of renewable energy created within the district, over a specified time period. \\
\addlinespace
\textbf{Electricity Consumption (E)} & The total amount of electricity consumed by the district over a year. \\
\bottomrule
\end{tabular}
\caption{Description of KPIs used in the CityLearn environment.}
\label{tab:kpi_description}
\end{table}
\clearpage
\newpage
\newpage
\vspace{10cm}
\section*{Appendix D}
\label{appendix_D}
This appendix outlines several technical components essential to the methodology and infrastructure behind our MARL experiments. Evaluating MARL algorithms at scale introduces significant computational and engineering challenges, requiring both efficient algorithmic implementations and scalable training pipelines. To address this, all algorithms evaluated in this study are implemented in JAX and trained using Sebulba, a scalable reinforcement learning architecture. These tools are critical for enabling reproducible, high-throughput experimentation and for leveraging distributed hardware such as GPUs and TPUs effectively.

In addition to implementation details, this appendix also covers supporting techniques and domain-specific tools used in the study, including the method used for quantifying agent importance and the Rainflow algorithm used to compute the DoD KPI. 

\subsection*{Scalable Reinforcement Learning Architectures}
JAX \citep{jax2018github} is a powerful numerical computing library for machine learning in Python, designed to support scalable ML tasks with features such as automatic differentiation, Single Instruction/Multiple Data (SIMD) programming, and hardware acceleration via Graphics Processing Units (GPUs) and Tensor Processing Units. JAX allows for efficient, large-scale computations through its use of just-in-time (JIT) compilation, vectorization, and parallelization. For RL, these capabilities are crucial to handle complex environments and policies, enabling rapid experimentation and scaling.

Scalable reinforcement learning architectures \citep{podracer}, such as Anakin and Sebulba, take full advantage of JAX's features to efficiently use TPUs.
\subsection*{Anakin Architecture}
The Anakin framework is designed for environments and agents that can both be implemented directly in JAX. It is an online learning system where the entire RL loop, comprising environment simulation, agent action selection, and learning updates, occurs directly on TPU devices. This architecture is highly efficient as it eliminates the need for frequent communication between the CPU host and the TPU cores, avoiding common bottlenecks such as data transfer delays.

\vspace{0.2cm}

Anakin leverages JAX primitives such as vmap and pmap to achieve large-scale parallelism and efficient hardware utilization. The vmap primitive enables automatic vectorization of the agent-environment interaction, transforming a single interaction into a batched version, where a batch of states can be processed in parallel, ensuring optimal utilization of each TPU core. 

\vspace{0.2cm}

Meanwhile, pmap enables data-parallel execution across multiple TPU cores, distributing the agent-environment interactions across all 8 cores within a single TPU device. By replicating this interaction loop across TPU cores, Anakin scales effectively across larger TPU slices or even entire TPU pods. Additionally, collective operations such as psum and pmean allow Anakin to aggregate results from different cores, ensuring that gradients are efficiently averaged for parameter updates across all participating cores. 

\vspace{0.2cm}

\newpage
\subsection*{Sebulba Architecture}
\begin{figure}
    \centering
    \includegraphics[width=0.5\linewidth]{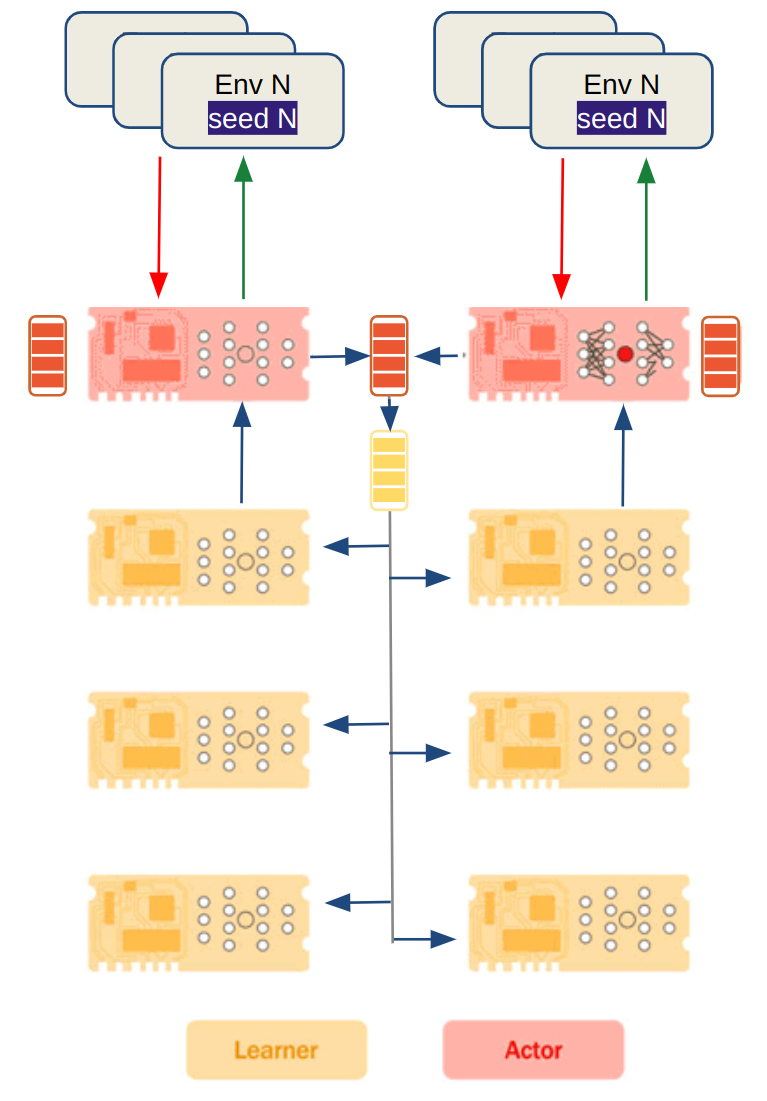}
  \caption{Sebulba podracer \cite{podracer} architecture diagram adapted from \cite{instadeepblog}.}
    \label{fig:Sebulba}
\end{figure}

Sebulba, illustrated by Figure \autoref{fig:Sebulba}, is a scalable architecture for RL that is designed to handle arbitrary environments, including those that cannot be compiled to run on TPUs. Unlike Anakin, which requires environments to be implemented in JAX, Sebulba introduces an actor-learner decomposition, which allows for efficient scaling while supporting diverse and complex environments such as Atari games or custom simulators running on CPUs.

\vspace{0.2cm}

In Sebulba, the computation is split between actors and learners. Actors interact with the environment to gather experience, while learners focus on training the model by updating its parameters. This separation allows for the efficient parallelization of both environment interactions and parameter updates, making the system more flexible and scalable.

\vspace{0.2cm}

The role of an actor is to execute the policy, gather experiences, and generate trajectories by interacting with the environment. Actors are responsible for sampling observations from the environment, selecting actions using the policy, and collecting the corresponding rewards and new states. The actors send these trajectories to the learners, who then perform gradient updates.

\vspace{0.2cm}

Meanwhile, Learners receive the trajectories from actors and use them to compute gradients and update the policy.The updated parameters are then sent back to the actors for further environment interactions.

\vspace{0.2cm}

Sebulba allows these actors and learners to operate asynchronously, meaning that learners can continue updating the policy while actors are collecting new trajectories. This setup helps to maximize utilization of both CPU and TPU resources.

\newpage
\subsection*{Rainflow Algorithm}
The Rainflow-counting algorithm is a method used in fatigue analysis to simplify complex stress or load histories by converting variable amplitude load cycles into a series of constant amplitude reversals. It systematically identifies and counts closed cycles in a load sequence, enabling the estimation of fatigue damage and component life. Developed by Tatsuo Endo and M. Matsuishi in 1974, the algorithm models the material memory effect seen in stress-strain hysteresis, facilitating more accurate predictions of structural fatigue under varying loads.

The rainflow cycle counting algorithm is summarized as follows:

\begin{enumerate}
    \item Rotate the loading history 90° such that the time axis is vertically downward and the load time history resembles a pagoda roof.
    
    \item Imagine a flow of rain starting at each successive extremum point.
    
    \item Define a loading reversal (half-cycle) by allowing each rainflow to continue to drip down these roofs until:
    \begin{enumerate} 
        \item It falls opposite a larger maximum (or smaller minimum) point.
        \item It meets a previous flow falling from above.
        \item It falls below the roof.
    \end{enumerate}
    
    \item Identify each hysteresis loop (cycle) by pairing up the same counted reversals.
\end{enumerate}
Consider the load profile presented in \autoref{fig:loadexample}, extracting the load profile from the SoC using this method will be as the following:
\begin{figure}
    \centering
    \includegraphics[width=0.5\linewidth]{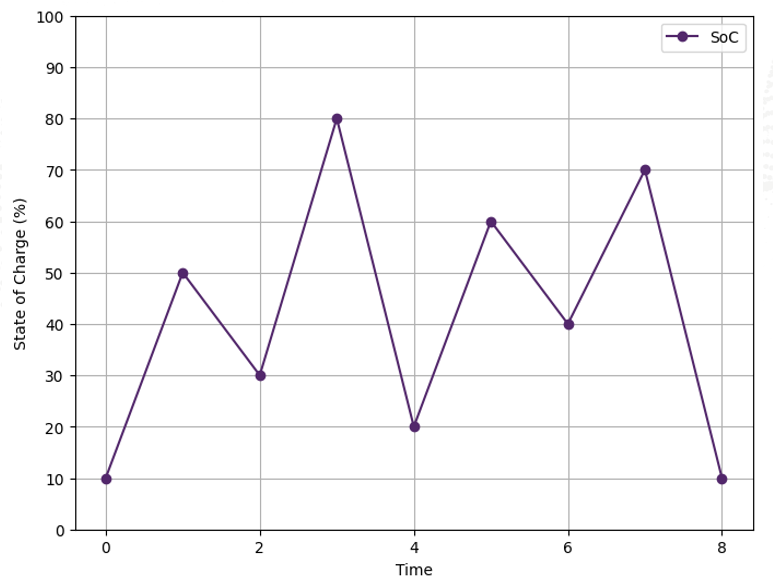}
    \caption{An example State of Charge (SoC) graph used to illustrate the application of the Rainflow Algorithm.}
    \label{fig:loadexample}
\end{figure}
\begin{enumerate}
    \item Rotate the load time history 90° clockwise. See 
    \autoref{fig:rotatedexample}
    \begin{figure}
    \centering
    \includegraphics[width=0.4\linewidth]{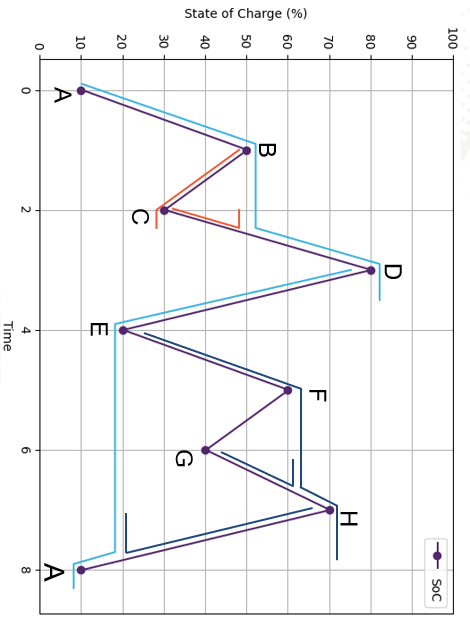}
    \caption{Rainflow Algorithm application example}
    \label{fig:rotatedexample}
\end{figure}
    \item Designate A as the first extremum point, the largest peak in this load time history.
    \item Identify the first largest reversal A–D as the flow of rain starts at A and falls off the second extremum point D, the smallest valley in this load time history.
    \item Identify the second largest reversal D–A as the flow initiates at D and ends at the other extremum point, which happens to be the first one, A.
    \item In the first largest reversal A–D,
    \begin{enumerate}
        \item Identify a reversal B–C as the rain starts flowing at B and terminates at C because D is a larger maximum than B.
        \item Identify a reversal C–B as the rain starts flowing at C and meets a previous flow at B.
        \item  Complete all the points in the first large reversal A–D.
    \end{enumerate}
    \item  In the second largest reversal D–A,
    \begin{enumerate}
        \item  Identify a reversal E–H as the rain starts flowing at E and falls off the roof at H.
        \item  Identify a reversal H–E as the rain starts flowing at point H and meets a previous flow at E.
        \item Identify a reversal F–G as the rain starts flowing at F and terminates at G because H is a larger maximum than F.
        \item Identify a reversal G–F as the rain starts flowing from the successive extremum point G and meets a previous flow at F.
        \item  Complete all the points in the second largest reversal D–A.
        
    \end{enumerate}
  \item  The rainflow cycle counting results in terms of reversals and cycles are given in \autoref{tab:reversal} and \autoref{tab:cycles}, respectively. 

\end{enumerate}

\begin{minipage}{0.45\textwidth}
    \centering
    
    \captionof{table}{Reversals Table}
    \label{tab:reversal}
    \begin{tabular}{c c c c c c}
        \toprule
        \textbf{From} & \textbf{To} & \textbf{From} & \textbf{To} & \textbf{Range} \\
        \midrule
        A & D & 10 & 80 & 70 \\
        D & A & 80 & 10 & 70 \\
        B & C & 50 & 30 & 20 \\
        C & B & 30 & 50 & 20 \\
        E & H & 20 & 70 & 50 \\
        H & E & 70 & 20 & 50 \\
        F & G & 60 & 40 & 20 \\
        G & F & 40 & 60 & 20 \\
        \bottomrule
    \end{tabular}
\end{minipage}%
\hfill
\begin{minipage}{0.45\textwidth}
    \centering
    
    \captionof{table}{Cycles Table}
    \label{tab:cycles}
    \begin{tabular}{c c c c c c}
        \toprule
        \textbf{From} & \textbf{To} & \textbf{From } & \textbf{To} & \textbf{Range} \\
        \midrule
        A & D & 10 & 80 & 70 \\
        B & C & 50 & 30 & 20 \\
        E & H & 20 & 70 & 50 \\
        F & G & 60 & 40 & 20 \\
        \bottomrule
    \end{tabular}
\end{minipage}

\vspace{0.2cm}

Using this algorithm the discharge cycles can be determined and the average DoD across these discharges is then computed.

\subsection*{Agent Importance}
Relying solely on aggregate performance metrics in MARL is insufficient, as these can mask significant performance heterogeneity. An algorithm's high average reward may be the product of a few "all-star" agents compensating for others that have learned sub-optimal policies. An analysis of individual agent contributions is therefore essential to understand the emergent coordination dynamics. This is critical for real-world applications, as it reveals whether an algorithm has learned a robust, collaborative solution or a brittle policy that marginalizes certain agents, thereby compromising system-wide reliability.

One way to compute the agents' contribution is through the use of the Shapley value \citet{Shapley1951}. Originating from game theory, the Shapley value addresses the issue of payoff distribution within a "grand coalition" (i.e., a cooperative game) and quantifies the contribution of each coalition member toward completing a task. Specifically, consider a cooperative game $\Gamma = (N, v)$, where $N$ is the set of all players, and $v$ is the payoff function used to measure the "profits" earned by a given coalition (or subset) $C \subseteq N \setminus \{i\}$. The marginal contribution of player $i$ is then given by:

\begin{equation}
    \varphi_i(C) = v(C \cup \{i\}) - v(C)
\end{equation}

The Shapley value of each player $i$ can then be computed as:

\begin{equation}
    S_i(\Gamma) = \sum_{C \subseteq N \setminus \{i\}} \frac{|C|!(|N| - |C| - 1)!}{|N|!} \cdot \varphi_i(C)
\end{equation}

\vspace{0.2cm}

Calculating Shapley values in the context of MARL presents two specific challenges: 
\begin{enumerate}
    \item It requires computing $2^n - 1$ possible coalitions for a potential $n$, which is computationally prohibitive.
    \item It strictly requires the use of a simulator where agents can be removed from the coalition, and the payoff for the same states can be evaluated for each coalition.
\end{enumerate}

Thus, following the approach in \citep{efficientagentcontribution}, Agent Importance values is used in this work, a computationally efficient approximation of the Shapley values.

The Agent Importance  values are computed as an average of difference rewards and used as an efficient estimate of the Shapley value. The computation is done over samples collected per step during evaluation, rather than per episode, without the need to resample coalitions, and are then aggregated over all evaluation timesteps. This approach greatly improves the sample efficiency in estimation. Concretely, the Agent Importance is given by

\begin{equation}
    S^{AI}_{i}(\Gamma) = \frac{1}{T} \sum_{t=1}^{T} \left( r_{t} - r_{t}^{-i} \right)
    \label{eq:agent_imp}
\end{equation}
where \( T \) is the number of timesteps in a full evaluation interval, \( r_t \) is the team reward (i.e., the reward of the grand coalition) at timestep \( t \), and \( r_{t}^{-i} \) is the team reward when agent \( i \) performs a no-op action.

\vspace{0.5cm}

Applying \autoref{eq:agent_imp} poses a technical challenge as it requires comparing rewards between agents based on the same exact environment state at a given timestep. In MARL, most simulators are not easily resettable and/or stateless, which makes measuring one reward and undoing that step and then measuring a second reward difficult. To overcome this limitation, \citet{efficientagentcontribution} adopt a simple solution outlined in \autoref{alg:agent_imp_alg}, where a copy of the environment is created for each agent to be able to compute the Agent Importance.

\begin{algorithm}[H]
\caption{Agent Importance}
\label{alg:agent_imp_alg}
\DontPrintSemicolon
\SetAlgoLined
\SetKwInOut{Input}{Input}\SetKwInOut{Output}{Output}
\setlength{\algomargin}{0pt}
\Input{evaluation timestep $t$, marginal contribution dictionary $marginal\_contribution$}
\BlankLine
$env\_copies \gets deepcopy(env, len(agents))$

$r_{t} \gets env.step(selected\_actions)$

\For{$i \leftarrow 0\ \textbf{to}\ len(agents)$}{
$actions\_with\_no\_op \gets disable\_actions(selected\_actions, i)$

$add\_to\_dict(marginal\_contribution, i, (r_{t} - r_{t}^{-i}))$
}

\end{algorithm}
\newpage
\section*{Appendix E}
\label{appendix_E}
To complement the primary findings presented in the main text, this appendix provides additional experimental results and visualizations. These include extended performance metrics, ablation studies, and supplementary comparisons across algorithms and configurations. The purpose of this section is to offer deeper insights into models behavior and robustness. While not central to the core analysis, these results reinforce the validity of our conclusions and highlight important nuances relevant for future research and deployment.
\subsection*{Additional Results}
\subsubsection*{MAPPO Variance}
\begin{figure}[H]
    \centering
    \begin{subfigure}[b]{0.48\textwidth}
        \centering
        \includegraphics[width=\linewidth]{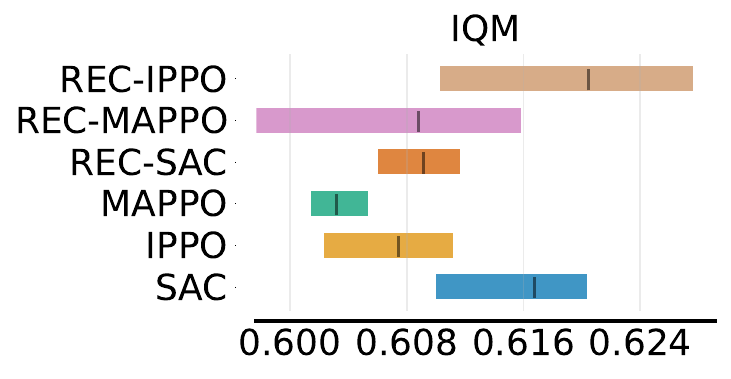}
        \caption{a) Best 3 Runs}
        \label{fig:bestthree}
    \end{subfigure}
    \hfill 
    \begin{subfigure}[b]{0.48\textwidth}
        \centering
        \includegraphics[width=\linewidth]{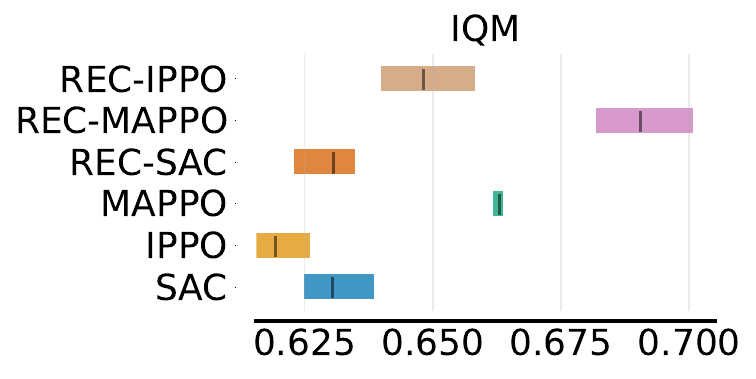}
        \caption{b) Worst 3 Runs}
        \label{fig:worstthree}
    \end{subfigure}

    \caption{
        \textbf{Performance Variance of MAPPO Algorithms in Best- vs. Worst-Case Runs.}
        This figure isolates the top three and bottom three performing runs for each algorithm to analyze their stability and performance consistency. Lower scores are better.
    }
    \label{fig:mappo_variance}
\end{figure}

The high variance that the MAPPO variants showcase can also be shown more evidently when looking at \autoref{fig:bestthree} and \autoref{fig:worstthree} where we can see that the MAPPO variants outperform their independent counterparts in top-tier runs but fall significantly behind in the worst-performing runs.

\subsubsection*{Close Performance across many KPIs}

\begin{figure}[H]
    \centering
    \begin{subfigure}[b]{0.33\textwidth}
        \centering
        \includegraphics[width=\linewidth]{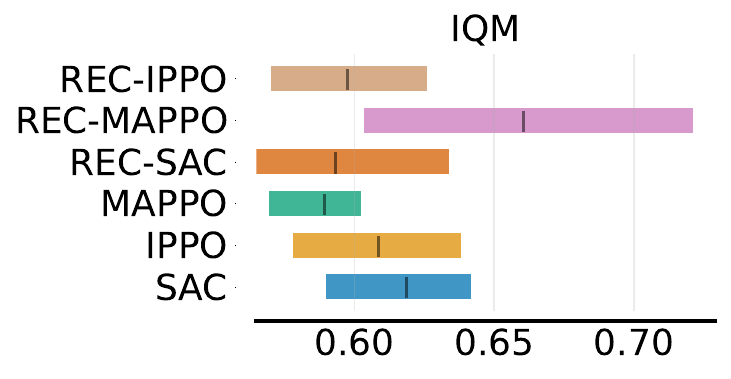}
        \caption{a) Unserved Energy}
        \label{fig:fig21}
    \end{subfigure}
    \hfill
    \begin{subfigure}[b]{0.32\textwidth}
        \centering
        \includegraphics[width=\linewidth]{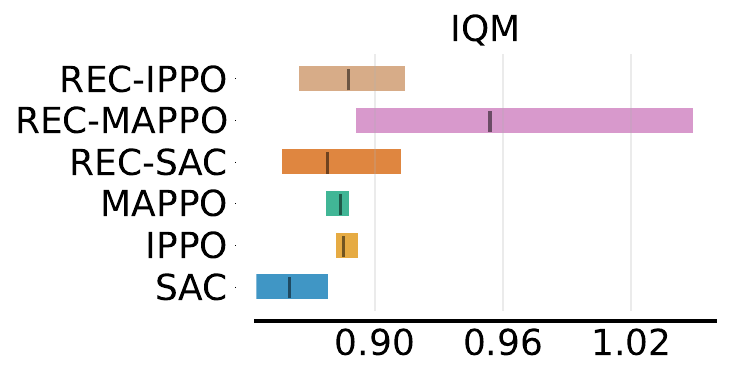}
        \caption{b) Annual Peak}
    \end{subfigure}
    \hfill
    \begin{subfigure}[b]{0.32\textwidth}
        \centering
        \includegraphics[width=\linewidth]{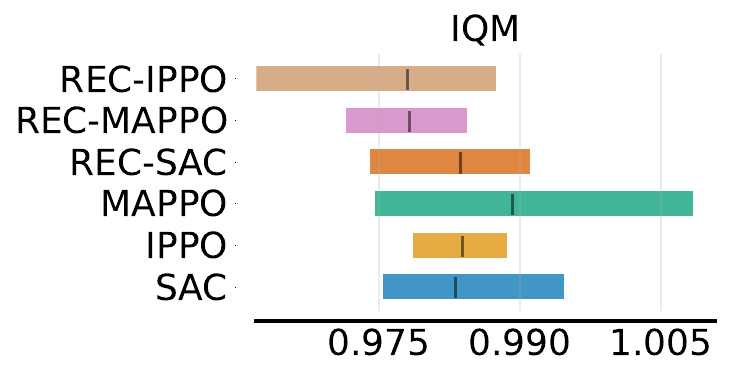}
        \caption{c) Daily Load Factor}
    \end{subfigure}

    \begin{subfigure}[b]{0.32\textwidth}
        \centering
        \includegraphics[width=\linewidth]{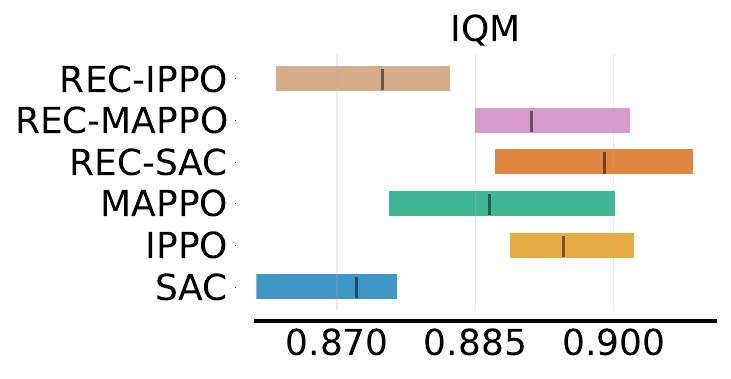}
        \caption{d) Daily Peak}
    \end{subfigure}
    \hfill
    \begin{subfigure}[b]{0.32\textwidth}
        \centering
        \includegraphics[width=\linewidth]{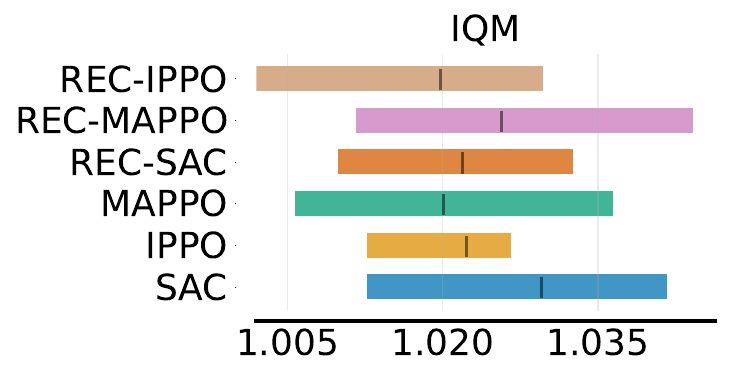}
        \caption{e) Monthly Load Factor}
    \end{subfigure}
    \hfill
    \begin{subfigure}[b]{0.32\textwidth}
        \centering
        \includegraphics[width=\linewidth]{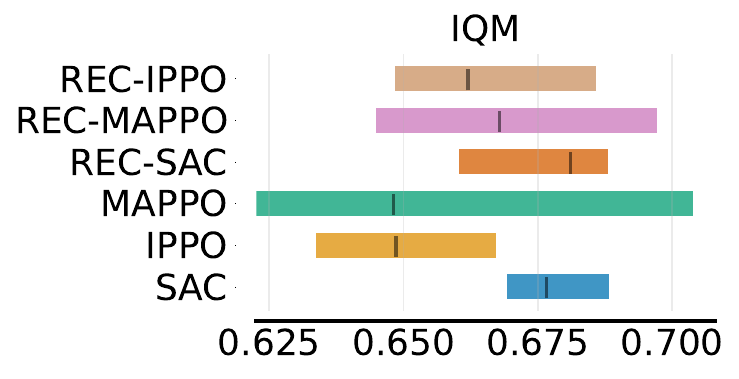}
        \caption{f) Thermal Resilience}
        \label{fig:26}
    \end{subfigure}

    \caption{
        \textbf{Performance Comparison on Secondary KPIs.}
        This figure presents a selection of key performance indicators where all tested algorithms demonstrate strong and broadly comparable results. Lower scores indicate better performance for all metrics. The overlapping confidence intervals and similar Interquartile Mean (IQM) scores across these plots indicate that no single algorithm holds a significant advantage for these objectives.
    }
    \label{fig:secondary_kpis}
\end{figure}

On several other KPIs, all algorithms achieve strong and comparable performance, as illustrated in the corresponding plots \autoref{fig:fig21}-\autoref{fig:26}.

\hfill

\subsubsection*{Tradeoff Analysis}
To gain a deeper understanding of model behavior beyond aggregate metrics, the trade-offs made across KPIs by each algorithm is compared. This analysis reveals how algorithms prioritize or sacrifice specific objectivesin pursuit of maximizing overall average performance. Understanding these trade-offs is essential for selecting models that align with specific priorities. To ensure meaningful comparison across KPIs with varying ranges, we apply min-max normalization using the global minimum and maximum observed values for each KPI across algorithms.

\begin{figure}[H]
    \centering
    \begin{subfigure}[b]{0.44\textwidth}
        \centering
        \includegraphics[width=\linewidth]{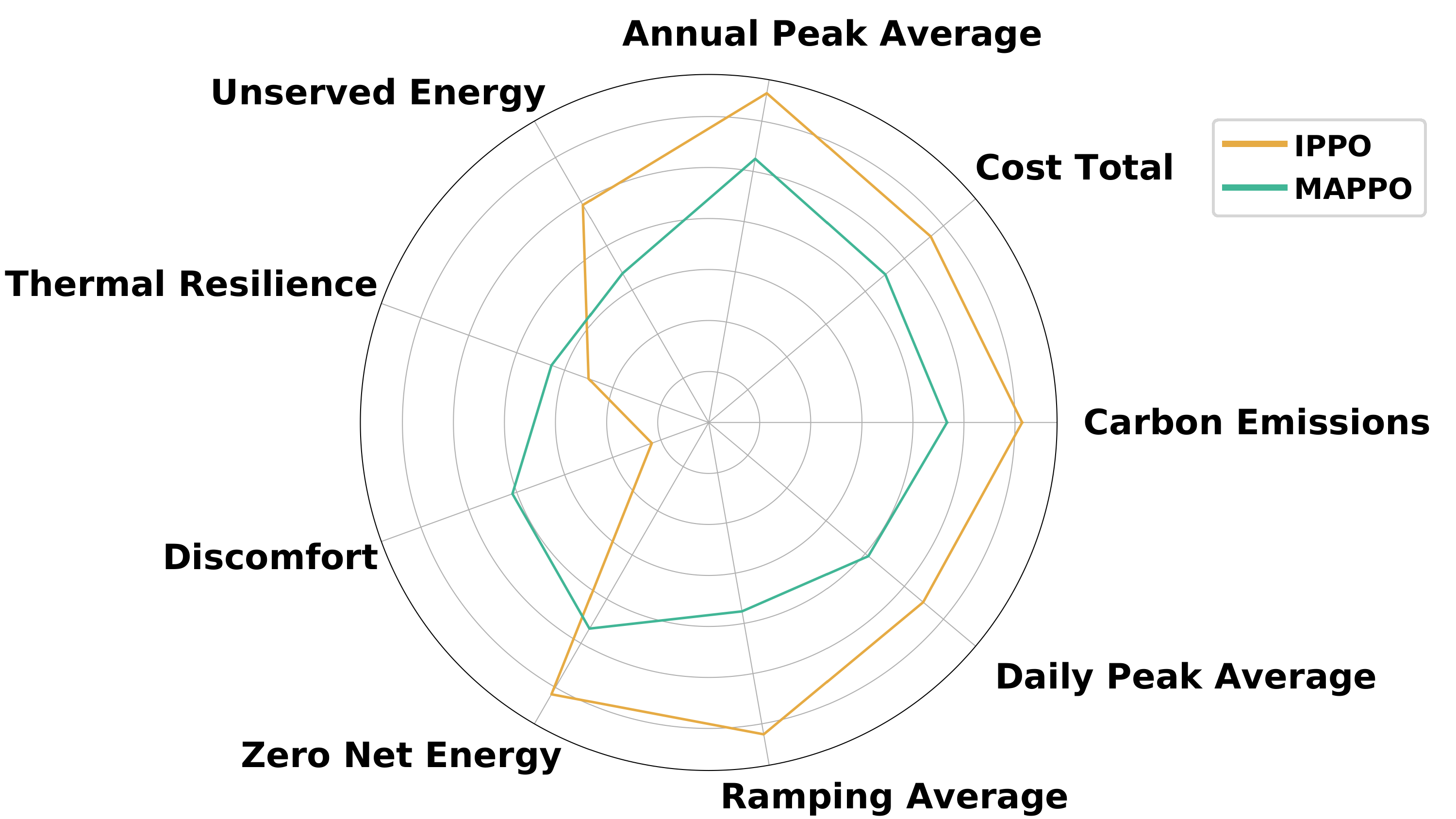}
        \caption{a) Spider Plot}
        \label{fig:spider_ippo_rec_mappo}
    \end{subfigure}
    \hfill
    \begin{subfigure}[b]{0.44\textwidth}
        \centering
        \includegraphics[width=\linewidth]{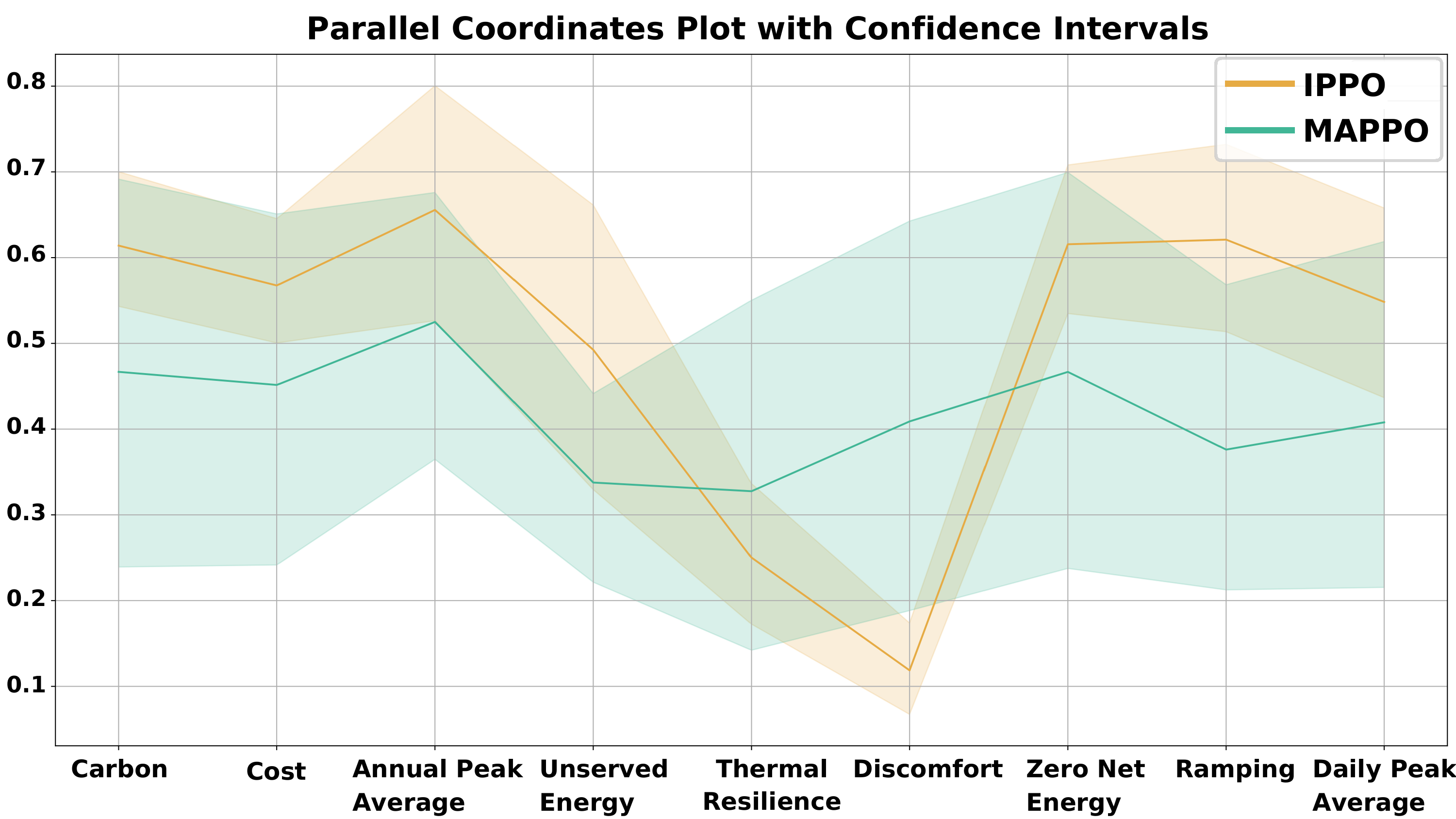}
        \caption{b) Parallel Coordinates Plot}
        \label{fig:parallel_ippo_mappo}
    \end{subfigure}
    \caption{
        \textbf{Trade-off Comparison between IPPO and MAPPO.}
        These plots visualize the performance of IPPO and MAPPO across multiple min-max normalized KPIs, where lower values are better. 
    }
    \label{fig:tradeoff_ippo_mappo}
\end{figure}

IPPO vs MAPPO. IPPO demonstrates a clear preference for minimizing KPIs like occupant discomfort and maximizing Thermal resilience often at the expense of KPIs such as Zero Net Energy and total cost. In contrast, MAPPO appears to pursue a more balanced approach, though this could be partially explained by its higher variability, occasionally achieving strong performance across all KPIs in some runs, and uniformly poor outcomes in others.

\begin{figure}[H]
    \centering
    \begin{subfigure}[b]{0.44\textwidth}
        \centering
        \includegraphics[width=\linewidth]{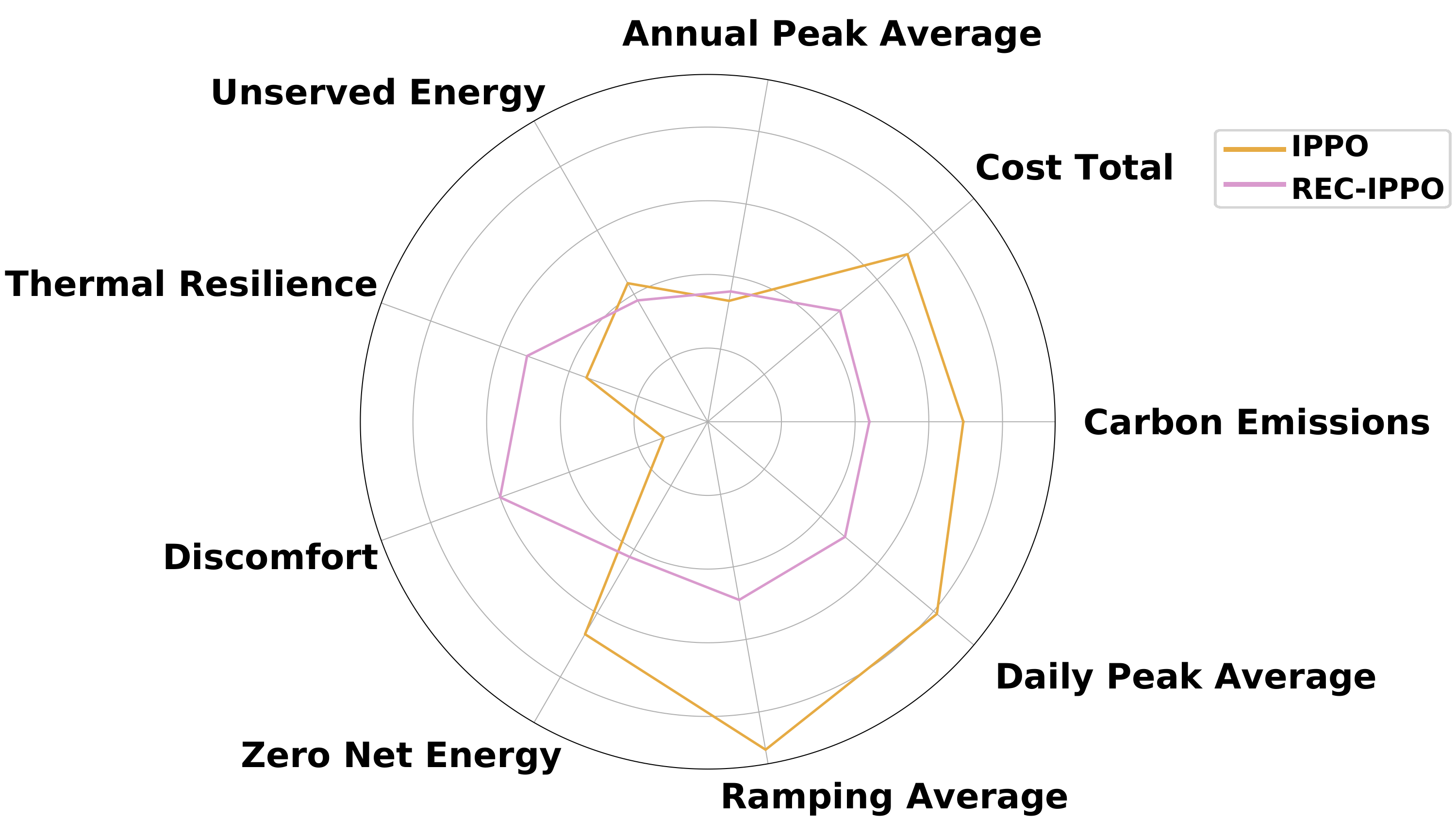}
        \caption{a) Spider Plot}
        \label{fig:spider_ippo_rec_ippo}
    \end{subfigure}
    \hfill
    \begin{subfigure}[b]{0.44\textwidth }
        \centering
        \includegraphics[width=\linewidth]{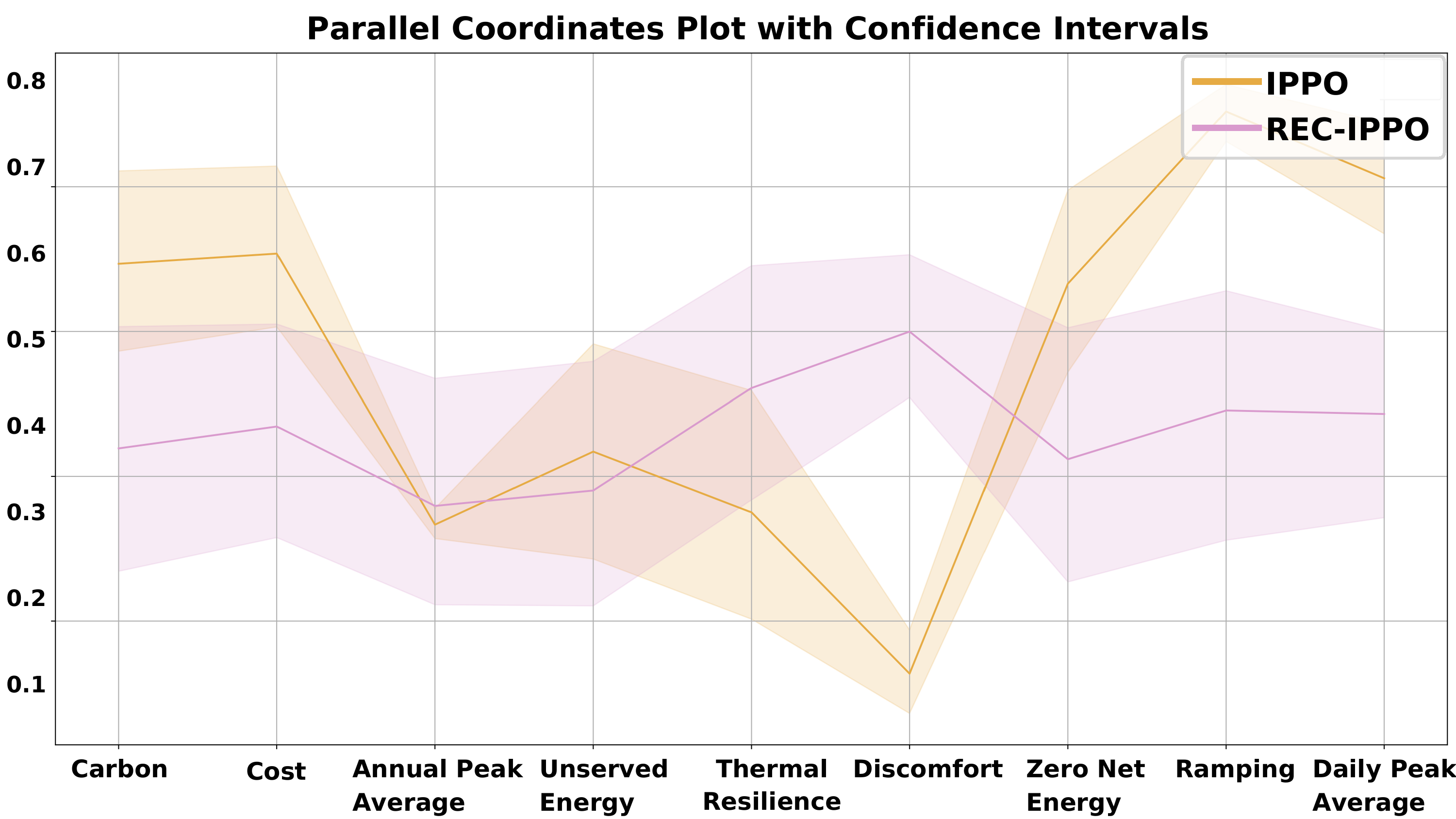}
        \caption{b) Parallel Coordinates Plot}
        \label{fig:parallel_ippo_rec_ippo}
    \end{subfigure}
    \caption{
        \textbf{Trade-off Comparison between IPPO and Rec-IPPO.}
        These plots compare the performance profiles of the feedforward and recurrent versions of IPPO across normalized KPIs where lower values are better. 
    }
    \label{fig:tradeoff_ippo_rec_ippo}
\end{figure}

IPPO vs Rec-IPPO. Temporal dependency leads to a sacrifice in comfort compared but makes notable gains in KPIs such as cost, ramping, and Zero Net Energy. This underscores the inherent trade-offs in the problem: improving performance in one domain often comes at a cost elsewhere, especially in more complex, memory-aware agents.

\begin{figure}[H]
    \centering
    \begin{subfigure}[b]{0.44\textwidth}
        \centering
        \includegraphics[width=\linewidth]{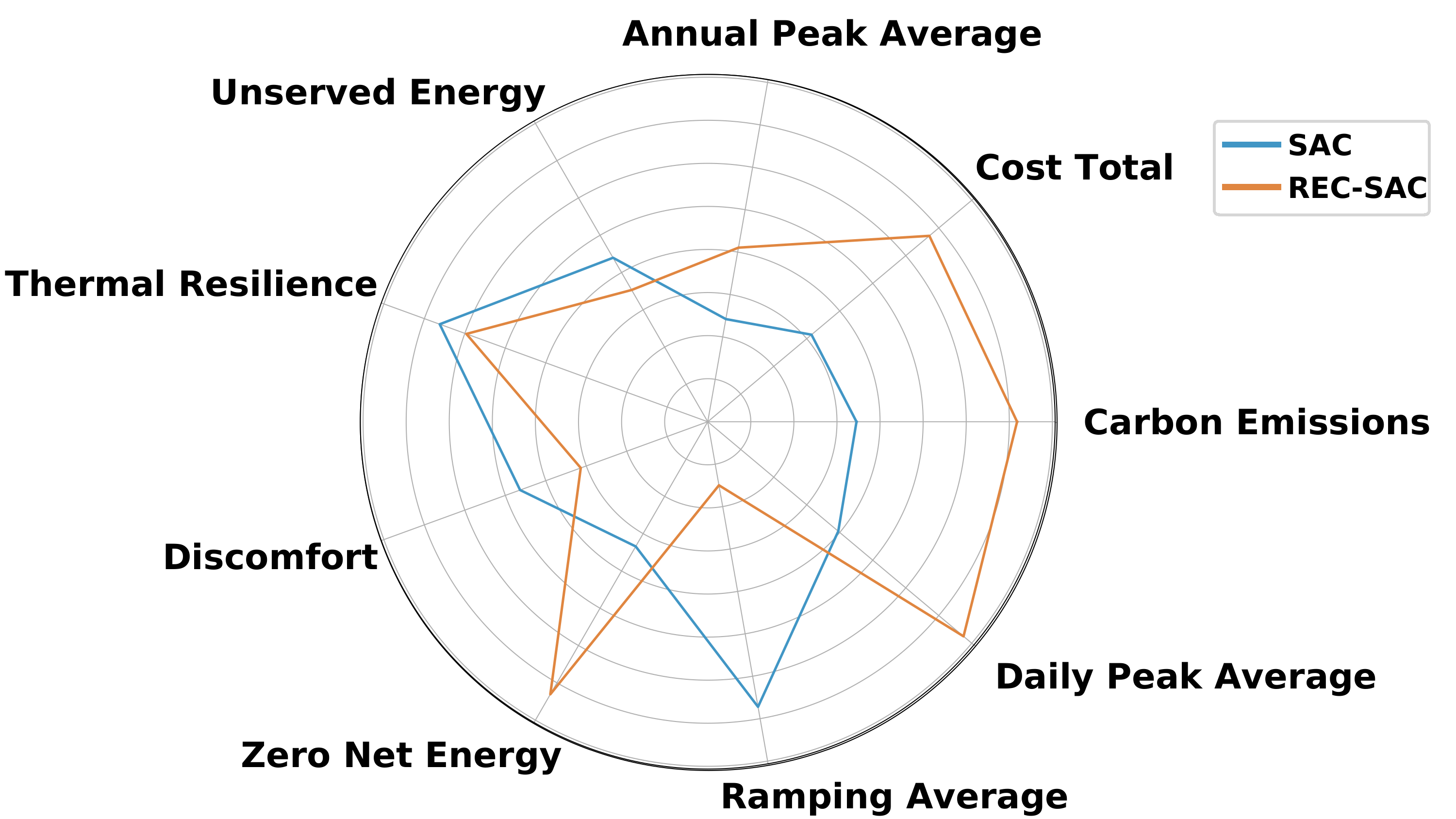}
        \caption{a) Spider Plot}
        \label{fig:spider_sac_recsac}
    \end{subfigure}
    \hfill
    \begin{subfigure}[b]{0.44\textwidth}
        \centering
        \includegraphics[width=\linewidth]{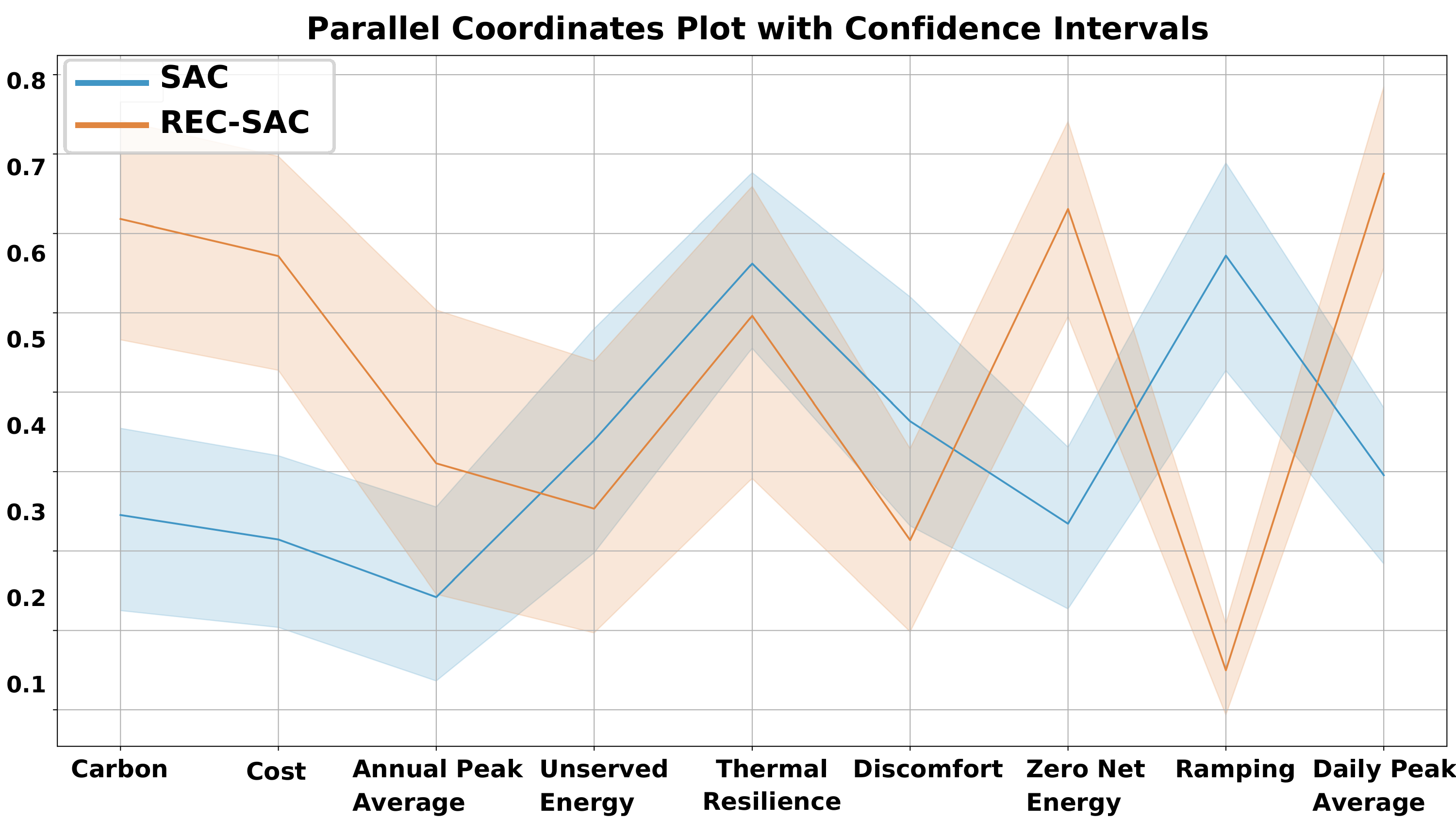}
        \caption{b) Parallel Coordinates Plot}
        \label{fig:parallsac_recsac}
    \end{subfigure}
    \caption{
        \textbf{Trade-off Comparison between SAC and Recurrent SAC (Rec-SAC).}
        These plots visualize the strategic differences between the standard and recurrent versions of the SAC algorithm across multiple normalized KPIs where lower values are better.
    }
    \label{fig:tradeoff_sac_recsac}
\end{figure}
\vspace{-0.2cm}
For SAC, temporal dependency improves the ramping KPI and discomfort at the cost of some lost performance on KPIs such as cost and Zero Net Energy.
\vspace{-0.4455cm}

\subsubsection*{Best performance across KPIs}
To assess the maximum potential of each algorithm, we analyze performance by selecting the highest value achieved for each KPI across the 10 runs. This approach highlights the peak capability of an algorithm with respect to specific metrics, offering insight into its best-case behavior.
For example, based on the values presented in \autoref{tab:table_X}, the highest observed performances across evaluations are 0.40, 0.45, 0.65, 0.35, and 0.60. These represent the upper bounds of the algorithm’s effectiveness under varying conditions, providing a useful benchmark for evaluating its performance ceiling.

\begin{minipage}{\textwidth}
    \centering
    
  \captionof{table}{Example of selecting the best observed performance per KPI across multiple evaluation runs.}
 \label{tab:table_X} 
\begin{tabular}{|c|c|c|c|c|c|}

\hline
\textbf{Run} & \textbf{Evaluation 1} & \textbf{Evaluation 2} & \textbf{Evaluation 3} & \textbf{Evaluation 4} & \textbf{Evaluation 5} \\ \hline
\textbf{Run 1} & 0.7 & 0.8  & 0.5  & 0.4  & 0.85 \\ \hline
\textbf{Run 2} & 0.6 & 0.7  & 0.55 & 0.45 & 0.75 \\ \hline
\textbf{Run 3} & 0.9 & 0.85 & 0.8  & 0.7  & 0.65 \\ \hline
\textbf{Run 4} & 0.5 & 0.6  & 0.55 & 0.35 & 0.4  \\ \hline
\textbf{Run 5} & 0.8 & 0.75 & 0.7  & 0.6  & 0.65 \\ \hline
\end{tabular}

\end{minipage}

\begin{figure}[H]
    \centering
    \begin{subfigure}[b]{0.32\textwidth}
        \centering
        \includegraphics[width=\linewidth]{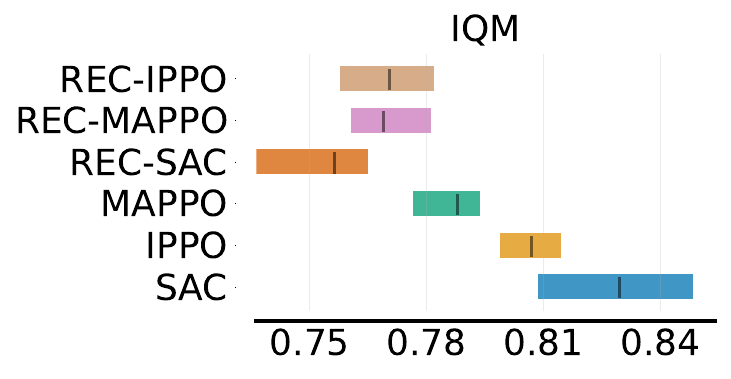}
        \caption{a) Carbon Emissions}
        \label{fig:carbon_emiss_best}
    \end{subfigure}
    \hfill
    \begin{subfigure}[b]{0.32\textwidth}
        \centering
        \includegraphics[width=\linewidth]{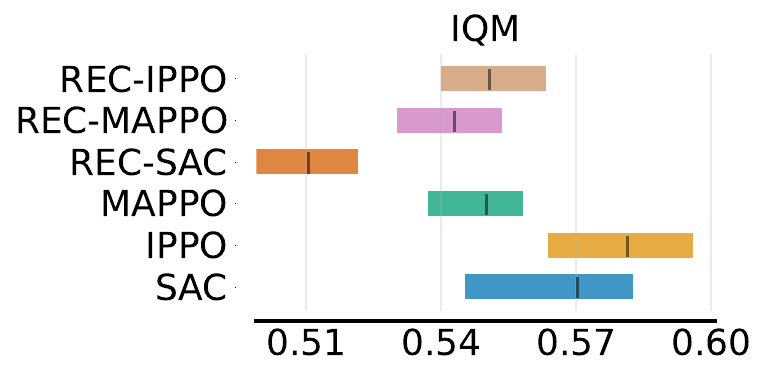}
        \caption{b) Unserved Energy}
        \label{fig:unserved_best}
    \end{subfigure}
    \hfill
    \begin{subfigure}[b]{0.32\textwidth}
        \centering
        \includegraphics[width=\linewidth]{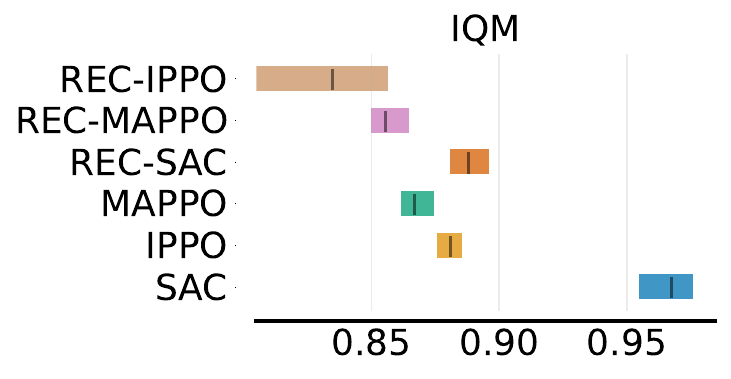}
        \caption{c) Ramping}
        \label{fig:ramping_best}
    \end{subfigure}

    \begin{subfigure}[b]{0.32\textwidth}
        \centering
        \includegraphics[width=\linewidth]{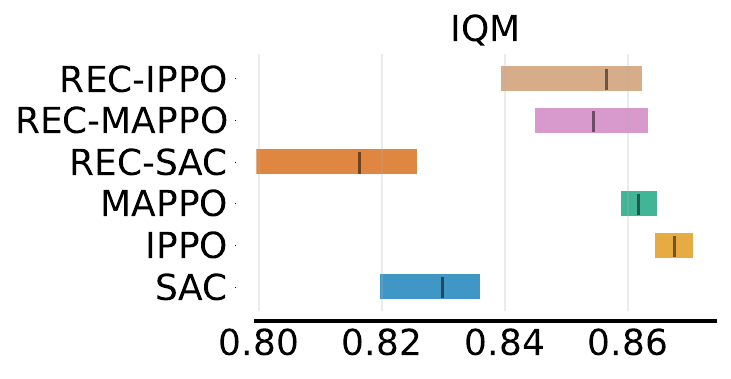}
        \caption{d) Annual Peak}
        \label{fig:peak_best}
    \end{subfigure}
    \hfill
    \begin{subfigure}[b]{0.32\textwidth}
        \centering
        \includegraphics[width=\linewidth]{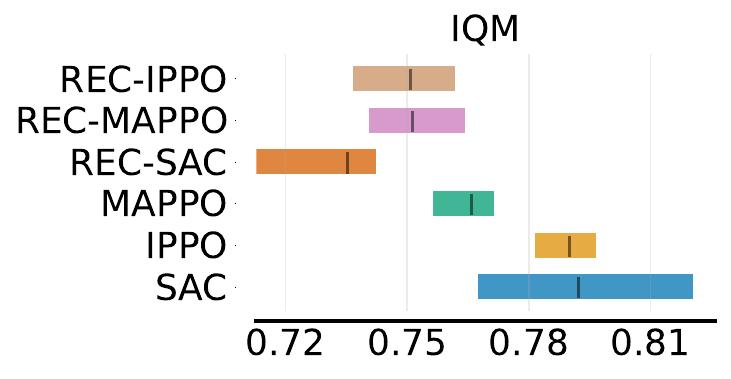}
        \caption{e) Total Cost}
        \label{fig:cost_best}
    \end{subfigure}
    \hfill
    \begin{subfigure}[b]{0.32\textwidth}
        \centering
        \includegraphics[width=\linewidth]{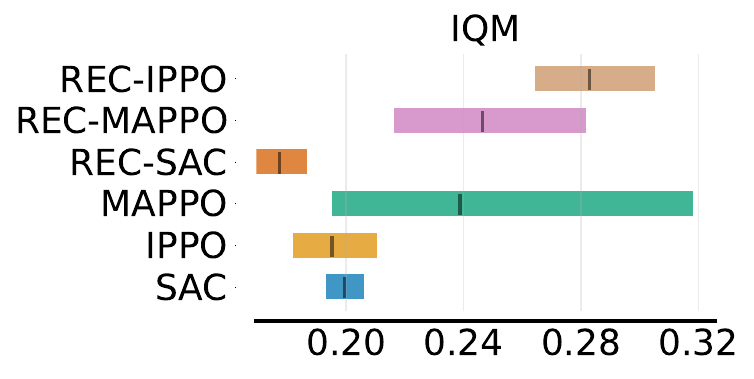}
        \caption{f) Discomfort Proportion}
        \label{fig:discomfort_best}
    \end{subfigure}

    \caption{
        \textbf{Best-Case Performance of Each Algorithm Across Key KPIs.}
        This figure highlights the peak performance achieved by each algorithm on six different metrics. This shows that recurrent models frequently achieve the best individual scores suggesting that they are capable of finding superior solutions for specific objectives contrasting with an algorithm like IPPO, which has a strong average score but may not reach the same performance peaks on individual KPIs.
    }
    \label{fig:best_kpi_performance}
\end{figure}

REC-SAC consistently achieves top performance across multiple KPIs, particularly excelling at minimizing discomfort proportion, where it even outperforms IPPO. Interestingly, temporal dependency consistently allows models to achieve the highest score. IPPO, though never best-in-class for any individual KPI, was shown to achieve the highest average score overall \autoref{fig:aggscore}, this might be largely due to its strength in discomfort minimization and a solid all-around performance. This suggests that while REC-SAC offers stronger performance peaks, IPPO exhibits superior consistency when balancing multiple objectives.

\subsubsection*{Scalability}
Scalability is a critical property in MARL, as many real-world applications involve systems where the number of agents is dynamic or significantly larger than those used during training. Adding or removing agents is common and retraining models for every configuration is impractical. As such, evaluating how well a policy generalizes when scaled to more agents than seen during training is essential for assessing the robustness and real-world applicability of MARL algorithms. In this experiment, we analyze performance when algorithms are trained with 3 agents and evaluated with 6, providing insight into their ability to adapt to increased coordination complexity and joint action space size.

\begin{figure}[H]
    \centering
    \includegraphics[width=0.5\linewidth]{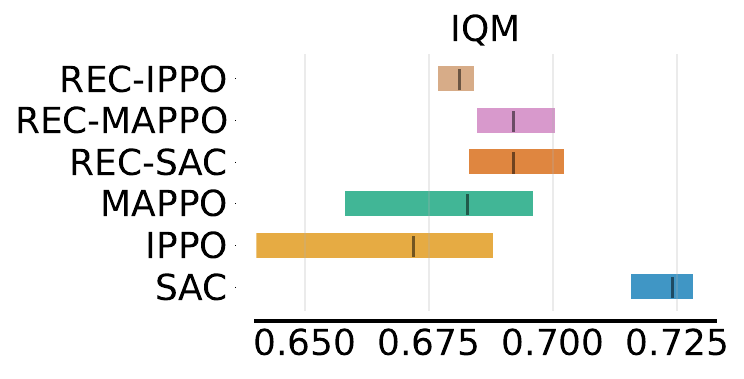}
    \caption{Performance of algorithms trained with 3 agents and tested on 6 agents}
    \label{fig:scalability}
\end{figure}

Most algorithms demonstrate robust scalability as the number of agents increases, see \autoref{fig:scalability}. Among them, IPPO consistently emerges as the top performer, indicating that its decentralized learning framework is particularly well-suited for multi-agent environments with increasing complexity. This scalability is a critical strength for real-world applications where system size can vary significantly. Interestingly, while REC-IPPO previously lagged behind leading algorithms, it now shows competitive performance.

\subsubsection*{Modified Observation}

For centralized training, the original observation space was formed by concatenating the observations of all agents into a single input. While this provides a comprehensive global view, it also introduces a significant amount of redundancy. Many agents receive overlapping or less crucial information, especially when predictive features like multi-step forecasts are involved. This redundancy can inject noise into the learning process and make the policy more sensitive to irrelevant features.

In this comparison, Rec-MAPPO is assessed across three different seeds under two configurations: the original observation space and a modified version where future predictions of outdoor temperature and electricity pricing (at 6/12/24 hours) were removed.

\begin{figure}[H]
    \centering
    \includegraphics[width=0.5\linewidth]{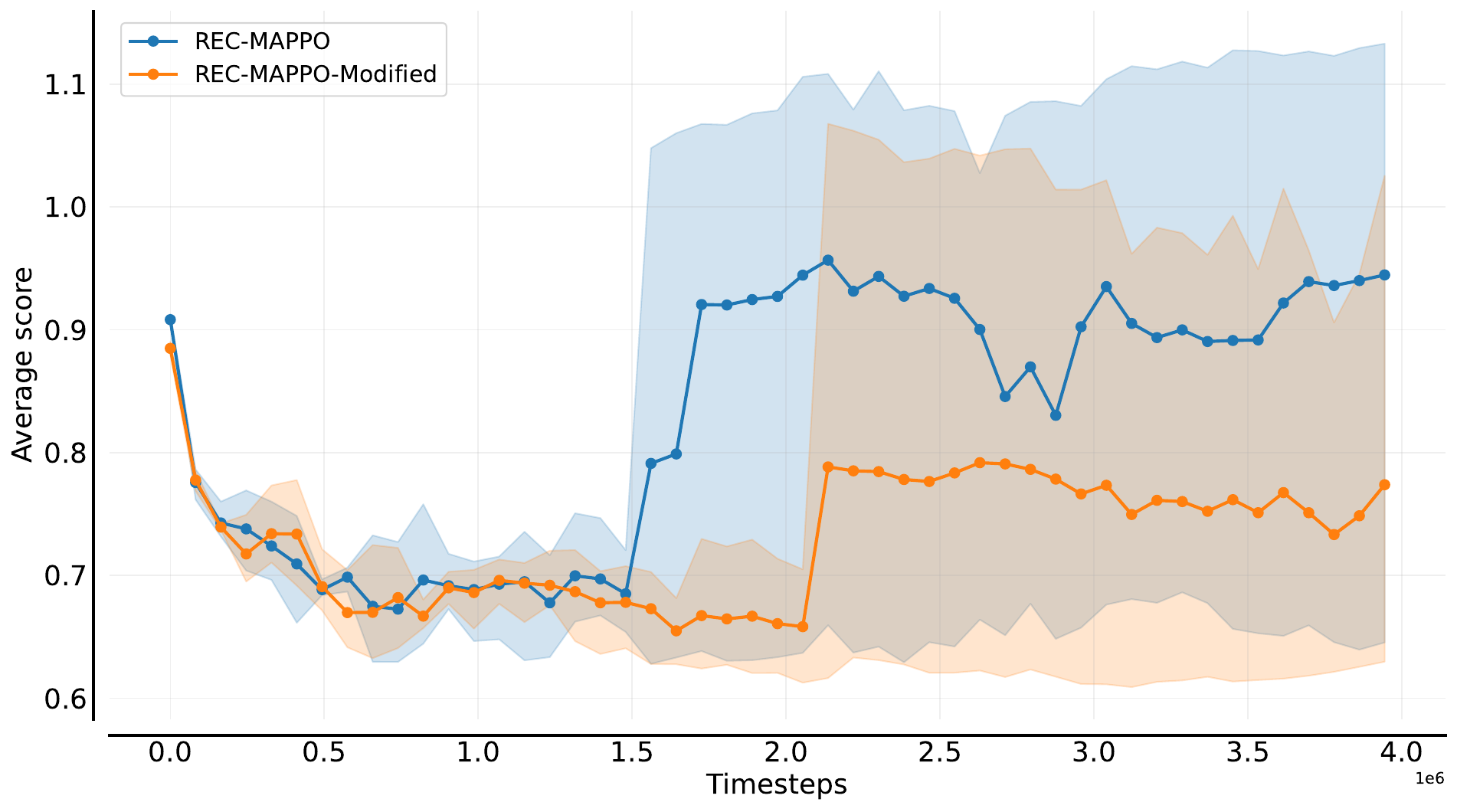}
    \caption{Sample efficiency curve of Rec-MAPPO for original observation vs modified observation}
    \label{fig:modified_obs}
\end{figure}

The results show that with the modified observation space, Rec-MAPPO achieves better average performance across the three seeds and exhibits less divergence between runs see \autoref{fig:modified_obs}. This implies that the removal of potentially noisy or irrelevant predictive information can lead to more stable and higher-performing models. In this case, simplifying the observation space has a positive effect on performance, reducing variance and improving reliability. These findings highlight the critical role of observation design, and suggest that further refinement of input observation could unlock even greater performance gains.
\end{document}